\documentclass[journal]{IEEEtran}
\usepackage{cite}
\usepackage[pdftex]{graphicx}
\DeclareGraphicsExtensions{.pdf,.jpeg,.png}
\usepackage{amsmath}
\usepackage{amssymb}
\usepackage{dsfont}
\usepackage{algorithmic}
\usepackage[caption=false,font=footnotesize]{subfig}
\usepackage{color}
\usepackage[table,xcdraw]{xcolor}

\newcommand{\set}[1]{\left\{#1\right\}}
\newcommand{\sbrk}[1]{\left[#1\right]}
\newcommand{\paren}[1]{\left(#1\right)}
\newcommand{\floor}[1]{\left\lfloor#1\right\rfloor}

\newcommand{\e}{\epsilon}
\newcommand{\half}{\frac{1}{2}}

\newcommand{\teq}{\triangleq}
\newcommand{\given}[2]{\left.#1\right|#2}
\newcommand{\mP}[1]{\mathbb{P}\left\{#1\right\}}
\newcommand{\cP}[2]{\mathbb{P}\left[\given{#1}{#2}\right]}

\newcommand{\figref}[1]{Fig. \ref{#1}}
\newcommand{\secref}[1]{Section \ref{#1}}
\newcommand{\algref}[1]{Algorithm \ref{#1}}
\newcommand{\lemref}[1]{Lemma \ref{#1}}
\newcommand{\thmref}[1]{Theorem \ref{#1}}

\newcommand{\tabref}[1]{Table \ref{#1}}
\newcommand{\lineref}[1]{line \ref{#1}}

\newcommand{\algorithmicnote}{\textbf{Note:}}
\newcommand{\NOTE}{\item[\algorithmicnote]}
\newcommand{\algorithmicinput}{\textbf{Input:}}
\newcommand{\INPUT}{\item[\algorithmicinput]}
\newcommand{\algorithmicoutput}{\textbf{Output:}}
\newcommand{\OUTPUT}{\item[\algorithmicoutput]}

\newtheorem{definition}{Definition}
\newtheorem{thm}{Theorem}
\newtheorem{prop}{Proposition}
\newtheorem{lem}{Lemma}
\newtheorem{assum}{Assumption}

\DeclareMathOperator*{\argmax}{arg\,max}

\newcounter{MYtempeqncnt}

\begin{document}

\title{Multi-user Communication Networks: \\A Coordinated Multi-armed Bandit Approach}
%
%
%

\author{Orly~Avner,~\IEEEmembership{Student Member,~IEEE,}
        and Shie~Mannor,~\IEEEmembership{Senior Member,~IEEE} %

\thanks{O. Avner and S. Mannor are with the Department of Electrical Engineering, Technion - Israel Institute of Technology, Haifa,
Israel, 32000.}
\thanks{Manuscript received August 08, 2018.}
}

\maketitle

\begin{abstract}
Communication networks shared by many users are a widespread challenge nowadays. In this paper we address several aspects of this challenge simultaneously: learning unknown stochastic network characteristics, sharing resources with other users while keeping coordination overhead to a minimum. The proposed solution combines Multi-Armed Bandit learning with a lightweight signalling-based coordination scheme, and ensures convergence to a stable allocation of resources.
Our work considers single-user level algorithms for two scenarios: an unknown fixed number of users, and a dynamic number of users. Analytic performance guarantees, proving convergence to stable marriage configurations, are presented for both setups. 
The algorithms are designed based on a system-wide perspective, rather than focusing on single user welfare. Thus, maximal resource utilization is ensured.
An extensive experimental analysis covers convergence to a stable configuration as well as reward maximization. Experiments are carried out over a wide range of setups, demonstrating the advantages of our approach over existing state-of-the-art methods. 
\end{abstract}

\section{Introduction}\label{sec:intro}
\IEEEPARstart{T}{he} world of modern multi-user communication networks poses many challenges that serve as an inspiration for our work.
We focus on distributed setups such as cognitive radio networks (CRNs) that consist of several users accessing a set of communication channels. The users' goal is to make the best possible use of network resources.

Achieving this goal is far from being simple in the setups we examine: channel characteristics are usually stochastic and initially unknown, and many users operate on an ``ad hoc'' basis that prevents communication and coordination between them. Moreover, the distributed nature of such networks prohibits any form of central control.

Solutions to this problem must therefore consist of an efficient user-level policy that incorporates learning and addresses the issue of multiple independent users targeting the same resources.

\subsection{Cognitive radio networks}
Cognitive radio networks, introduced in \cite{Mitola1999}, are a conceptual framework for modern communication networks that has gathered considerable interest in recent years. The main idea this framework proposes is that radios with enhanced capabilities can utilize communication resources better than traditional radios.
The issue of resource utilization is of great interest due to the shortage in available frequency ranges, combined with poor utilization of those frequency ranges that are allocated.

Radios with enhanced capabilities such as spectrum sensing, memory and computational power can identify and use ``gaps'' in transmissions of licensed traditional radios, thus increasing utilization. From an algorithmic point of view, this framework gives rise to several interesting questions due to its dynamic, stochastic, distributed nature. Over the last decade this challenging assortment of problems has gained considerable attention from researchers and engineers \cite{Arslan2007,Haykin2005}. Both theoretical and practical issues have been addressed, along with the necessary increase of regulatory support \cite{Regulation2013}.

\subsection{Multi-armed bandits}
Multi-armed bandits (MABs) are a well-studied framework from the world of machine learning. They model a sequential decision making problem in which a user repeatedly chooses one of $K$ actions in order to maximize her acquired reward. The characteristics of the actions (also known as arms) are initially unknown, and learning to identify the best action needs to be balanced with reward maximization, in what is known as the exploration-exploitation dilemma.
MABs have attracted much interest due to the wide range of applications they capture, combined with their relative simplicity, from both algorithmic and analytic points of view. Several papers propose solutions for the stochastic MAB problem \cite{Auer2002a,Auer2010,Garivier2011} in which rewards are drawn from some distribution, and for its adversarial counterpart \cite{Auer2002b}. However, they all focus on single users. The few papers dealing with multiple users usually assume a setting with some form of collaboration \cite{Awerbuch2008,Cesa-Bianchi2016}.

\subsection{The CRN-MAB framework}
Using MABs to model CRNs was first suggested in \cite{Jouini2010}, in a rather straightforward manner -- the channels of a communication network simply correspond to the arms of a MAB. An extension that also takes operational constraints into account appears in \cite{Avner2011}.

The multi-user scenario of the CRN-MAB setup was first introduced in \cite{Liu2010} and further explored in \cite{Anandkumar2011}. Both focused mainly on a fixed, known number of users. An algorithm that did not require any knowledge of the number of users and supported user arrival and departure was proposed in \cite{Avner2014}, and other approaches were introduced in \cite{Rosenski2015,Besson2017}.

However, this entire body of work assumes channel characteristics to be identical for all users. In reality, this may prove to be an unreasonable assumption. Users who are geographically far apart may experience different disturbances, leading to different channel reward distributions.

The work closest in spirit to ours is presented in \cite{Kalathil2014}, taking multi-user MABs with different channel characteristics into account. However, the authors incorporate the Bertsekas auction algorithm, that requires frequent information exchange, into their solution. Such an approach requires users to have distinct i.d.'s and knowledge of the number of users. This rather technical requirement hinders the ability of the algorithm to deal with a variable number of users, in addition to the price of communication itself. We compare our solution with this approach in detail in \secref{sec:experiments}. An extension of this paper appears in \cite{Nayyar2016}.

\subsection{Assignment problems}
Another way to approach the problem of multiple users with different rewards is through graph theory. In this setup users correspond to agents and channels to tasks, with rewards being the complementary of graph edge costs. The result is simply an assignment problem, i.e., maximum weight matching in a weighted bi-partite graph.

Several papers have tackled the assignment problem, but to the best of our knowledge none of them have proposed a solution that solves the multi-user CRN-MAB described above.

The classical Hungarian method \cite{Kuhn1955} requires some form of central control and assumes channel characteristics are known. The Bertsekas auction algorithm \cite{Bertsekas1988} is suitable for a distributed setting but requires direct, frequent communication between agents. The well-known Gale-Shapley algorithm \cite{Gale1962} converges to a stable marriage configuration, but does not take learning into account. In \cite{Leshem2012,Cohen2013} the Gale-Shapley algorithm was indeed applied to the CRN setup, without considering the need to learn channel characteristics.

Another work on distributed stable marriage considers a variant of the Gale-Shapley algorithm \cite{Floreen2010}. The paper itself is rather unrelated to our problem, but the potential function introduced by the authors was helpful in our analysis.

Another paper that aims at limiting information exchange between users, but does not address learning, is \cite{Amira2010}. Finally, the authors of \cite{Kipnis2009} and \cite{Gonczarowski2014} derive lower bounds on the information exchange required to solve assignment and stable marriage problems.

\subsection{Our contribution}
The novelty in our work stems from the combination of multiple users, different reward distributions, an unknown and possibly dynamic number of users and minimal coordination that does not require direct communication between users.

Allowing different reward distributions and limiting information exchange are a challenging combination. As explained in detail in \secref{ssec:performance}, reward maximization cannot be guaranteed in this setting. Instead, we focus on convergence to a stable configuration. Once users settle into a stable configuration, they can focus on resource utilization. When the transmitted content is in the form of streams (e.g., audio and video), maintaining the same communication channel between time slots is doubly important.

Our contribution consists of:
\begin{itemize}
	\item A coordination scheme for minimizing collisions during exploration and exploitation in multi-user communication networks
	\item An algorithm that combines learning and coordination, ensuring convergence to a stable configuration
	\item Theoretical guarantee of convergence in a flexbile scenario that consists of an unknown number of users with different reward distributions
	\item Extensive experiments demonstrating the algorithm's performance
\end{itemize}

The structure of this paper is as follows: \secref{sec:model} introduces the model and assumptions, together with a basic formulation. \secref{sec:CSM-MAB} and \secref{sec:static-analysis} describe the CSM-MAB algorithm, for the scenario of a fixed number of users, with its theoretical analysis. \secref{sec:D-CSM-MAB} and \secref{sec:dynamic-analysis} handle a dynamic number of users. Extensive experiments complement the theoretical analyses in \secref{sec:experiments}, and a discussion follows in \secref{sec:conclusion}.


\section{Model and Formulation}\label{sec:model}
In this section we describe our model and the assumptions we make in developing the algorithms and analyses, together with a mathematical formulation of the concepts used throughout the paper.

\subsection{System}\label{ssec:system}
The communication system we are dealing with consists of $K$ channels. A single user transmitting in a certain channel acquires a reward, corresponding to any chosen performance measure in real-life systems: throughput, bit rate, etc. We assume rewards are stochastic, bounded in the interval $\sbrk{0,1}$.
As is customary in ad-hoc networks, there is no central control.

\subsection{Users}\label{ssec:users}
Our users are a group of $N$ independent agents. Each agent observes only her own rewards, and gathers statistics concerning her own actions. She does not know the number of users sharing the network and in the dynamic setting she may join or leave the network at random points in time. In addition, users cannot exchange information directly.

We adopt a reward model that depends on both the identity of the user and the index of the channel, in order to reflect differences in the physical environment of users. Formally, a user $n$ sampling a channel $k$ will receive a reward drawn i.i.d. from a distribution with an expected value of $\mu_{n,k}$.

The notion of shared resources is modeled by the users' playing \emph{the same} bandit. When two or more users transmit in the same channel, they experience a collision. In this paper, collisions result in reward loss for all colliding users for the relevant time slot.

Throughout the paper the term \emph{configuration} refers to a mapping of users to channels.

\subsection{Performance measures}\label{ssec:performance}
In our work we adopt the point of view of a network designer, whose goal is to maximize resource utilization. We are therefore interested in system-wide performance measures, rather than user level ones.

A common performance measure for the multi-user CRN-MAB setup is the system-wide reward (see e.g., \cite{Anandkumar2011,Avner2014,Kalathil2014}). However, in the setup we are dealing with, in which the reward distribution varies from user to user, achieving the optimum in terms of system-wide reward requires frequent information exchange between users. Let us assume channel $k$ is the optimal channel for two users, $n_1$ and $n_2$. However, $\mu_{n_1,k}>>\mu_{n_2,k}$ and the next best channel for user $n_2$ yields a reward very similar to $\mu_{n_2,k}$. As a result, in the optimal configuration user $n_1$ must ``win'' the right to transmit in channel $k$. In order for user $n_2$ to agree to step down, she must receive explicit information from user $n_1$ regarding that user's reward in channel $k$. Incorporating learning into the process results in evolving preferences, further increasing the necessary information exchange.

We would like to avoid such heavy communication, for two reasons. First, communication between users may increase a network's vulnerability to attacks. In addition, implementing such protocols requires considerable resources, as we demonstrate in our experiments in \secref{sec:experiments}.

Our solution adopts a different performance measure: system-wide stability, in the Stable Marriage (SM) sense.

\begin{definition}
  A Stable Marriage Configuration (SMC) is an assignment of users to channels such that no two users would be willing to swap channels, had they known the true values of the expected rewards.
  Formally, for a pair of users $n_1$ and $n_2$:
  \begin{align*}
    C_1 &\triangleq \mathds{1}\set{\mu_{n_1,a_{n_1}}<\mu_{n_1,a_{n_2}}} \\
    C_2 &\triangleq \mathds{1}\set{\mu_{n_2,a_{n_2}}\leq\mu_{n_2,a_{n_1}}},
  \end{align*}
  where $a_{n_i}$ is the channel user $i$ is currently sampling.

  In an SMC,
  \begin{align*}
    C_1 \cdot C_2 = 0 \quad \forall n_1,n_2.
  \end{align*}
\end{definition}

Focusing on this performance measure ensures efficient use of system resources and does not require frequent or excessively informative communication.

\subsection{Limited coordination}\label{ssec:limit-coord}
The coordination protocol described in \secref{ssec:protocol} relies on two very simple building blocks, that are in line with common capabilities of CRNs.
Users are allowed to transmit in a single channel in each time slot, and observe the acquired reward.
In addition, they may sense all channels at once and observe a vector of binary signals, where ``1'' corresponds to a transmission in the channel and ``0'' corresponds to the lack of a transmission. This capacity can be viewed as a form of wideband spectrum sensing which is common in CRNs \cite{Sun2013}.
\algref{alg:CSM-MAB} relies on a combination of sensing and transmission according to a carefully planned temporal structure, which we call a super-frame (see \figref{fig:frames}). By following this protocol users can coordinate without directly communicating.

\subsection{Goal}\label{ssec:goal}
Our goal in this paper is to design efficient algorithms for the single user level, for both fixed and dynamic numbers of users.
Our solution will ensure convergence to an orthogonal (i.e., no more than one user per channel) stable marriage configuration in finite time, using only the limited form of coordination made possible by the actions described in \secref{ssec:limit-coord}. 

\section{CSM-MAB algorithm}\label{sec:CSM-MAB}
In this section we present the Coordinated Stable Marriage for Multi-Armed Bandits (CSM-MAB) algorithm (\figref{alg:CSM-MAB}). This algorithm assumes a fixed, yet unknown, number of users.

\subsection{Algorithm outline}\label{ssec:alg-outline}
Our algorithm combines a coordination protocol and a learning process.
It is a user-level algorithm for a fully distributed system, designed in order to achieve stability and orthogonality, as described in \secref{ssec:goal}.

\begin{figure}[!t]
	\centering
	\includegraphics[width=\columnwidth]{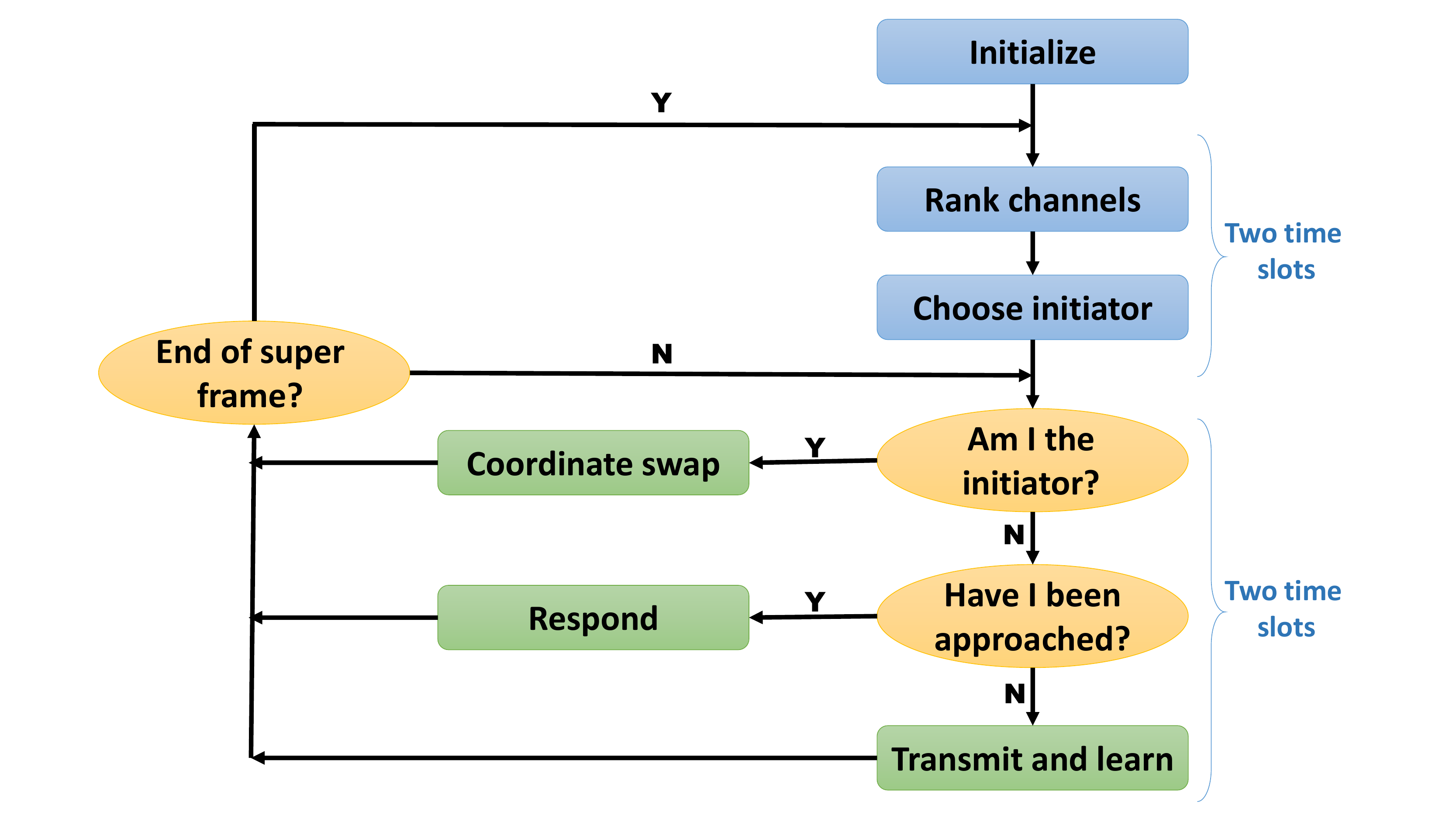}
	\caption{Flow of CSM-MAB algorithm}
	\label{fig:outline}
\end{figure}

After a short initialization phase, the algorithm follows the flow described in \figref{fig:outline}.
At the beginning of each super-frame users update their channel preferences and an initiator is determined. During the rest of the super-frame, a channel swap between  users is coordinated. Users not participating in the coordination process continue to transmit and learn during these time slots.
A detailed explanation of the coordination protocol and frame structure appears in the next section.

\begin{figure}
  \begin{algorithmic}[1]
  	\STATE \textbf{r}=0,\textbf{s}=0, pref=0
    \STATE \textbf{apply\_CFL}
    \FORALL{time slots $t$}
        \IF {Beginning of super-frame}
            \STATE \textbf{list}, \textbf{I} $\leftarrow$ \textbf{rank\_channels(r,s)}
            \STATE \textbf{choose\_initiator(list)}
            \STATE pref $\leftarrow$ 1
        \ELSE
            \IF {$n$ is the initiator}
                \STATE \textbf{coordinate\_swap}(\textbf{list},pref)
            \ELSIF {$n$ was approached by initiator}
                \STATE \textbf{respond(I)}
            \ELSE
                \STATE \textbf{r},\textbf{s} $\leftarrow$ \textbf{transmit\_and\_learn}
            \ENDIF
        \ENDIF
    \ENDFOR
  \end{algorithmic}
\caption{The CSM-MAB algorithm for user $n$}
\label{alg:CSM-MAB}
\end{figure}

The algorithm begins with a start-up phase, during which the CFL algorithm introduced in \cite{Leith2012} is applied. The \textbf{apply\_CFL} function yields an initial orthogonal configuration, so that each user starts off in a different channel.
After this phase, the body of the algorithm is executed. Each time a super-frame begins, all users execute the \textbf{rank\_channels} routine, which receives a user's local knowledge as input, and outputs a list of preferences over all channels (see \figref{alg:rank-channels}).
Users' local knowledge consists of the sum of rewards acquired so far from each channel ($\textbf{r}$), along with the number of times each channel has been sampled ($\textbf{s}$). The channels are ranked according to the Upper Confidence Bound (UCB) index \cite{Auer2002a}.

\begin{figure}
  \begin{algorithmic}[1]
    \INPUT \textbf{r,s}
    \FOR{$k\in\set{1,\ldots,K}$}
        \STATE $I_k = \frac{r_k}{s_k} + \sqrt{\frac{2\log t}{s_k}}$
    \ENDFOR
    \STATE list $\gets$ \textbf{sort\_descend}$\paren{\textbf{I}}$
    \OUTPUT list, \textbf{I}
  \end{algorithmic}
\caption{The \textbf{rank\_channels} routine}
\label{alg:rank-channels}
\end{figure}

The next step is choosing an initiator for the super-frame, according to the \textbf{choose\_initiator} subroutine (\figref{alg:choose-init}). All users who are not currently sampling the channel that maximizes their UCB index, raise a flag with some constant probability $\e$. If exactly one user raises such a flag, then she is the initiator for the super-frame. Otherwise, there is no initiator.
The value of the parameter $\e$ is chosen so that it maximizes the probability that exactly one user raises a flag, assuming all users would like to be initiators, as explained in further detail in \secref{app:a1-init}.

\begin{figure}
  \begin{algorithmic}[1]
    \INPUT list
    \IF [User seeks to change channel] {$list \neq \phi$}
        \STATE $flag_n\gets$ \textbf{rand}$\paren{\text{Bernoulli},\epsilon}$ \COMMENT{Raise flag w.p. $\e$}
        \IF {$\paren{flag_n = 1} \land \paren{flag_i = 0 \; \forall i \ne n}$}
            \STATE $initiator = n$ \COMMENT{User $n$ is initiator for this SF}
        \ELSIF {$\paren{flag_i = 0 \; \forall i} \lor \paren{\text{nnz}\paren{flag} > 1}$}
            \STATE $initiator = 0$ \COMMENT{No initiator for this SF}
        \ENDIF
    \ENDIF
  \end{algorithmic}
\caption{The \textbf{choose\_initiator} routine}
\label{alg:choose-init}
\end{figure}

Once the initiator has been selected, she iterates over her list of preferences. In order to upgrade to a certain channel, she transmits in it. The user currently occupying that channel senses the transmission, and decides whether she accepts the offer to swap. This decision is made based on the local knowledge and preferences of the responder: if swapping will improve upon her current choice of channel in terms of the UCB index (or, at least, will not worsen her situation), then she will accept. If the responder accepts, a swap takes places and the initiator does not negotiate any more swaps during the super-frame. Otherwise, the initiator updates her preference pointer, and in the next time slot she will attempt to swap to the next best option on her list of preferences. This process repeats itself until the initiator manages to coordinate a swap or she reaches the end of her preference list. The actions taken by the initiator are described in  \textbf{coordinate\_swap} (\figref{alg:coord-swap}), and the responder's actions appear in \textbf{respond} (\figref{alg:resp}).

\begin{figure}
  \begin{algorithmic}[1]
    \INPUT list, pref
    \IF [list not exhausted]{$pref > 0 \land list\paren{pref} \neq \bot$}
				\STATE \textbf{transmit}$\paren{list\paren{pref}}$
                \STATE $response \gets$ \textbf{sense}$\paren{list\paren{pref}}$
                \IF [Responder agreed or channel is available]{$response = 1$}
                    \STATE $a\paren{t} \gets$ \textbf{swap}$\paren{a_n\paren{t},list\paren{pref}}$
                    \STATE $pref \gets 0$
                \ELSE
                    \STATE $pref \gets pref + 1$ \COMMENT{Move to next best channel}
                \ENDIF
            \ENDIF
  \end{algorithmic}
\caption{The \textbf{coordinate\_swap} routine, performed by the initiator each mini-frame}
\label{alg:coord-swap}
\end{figure}

\begin{figure}
  \begin{algorithmic}[1]
    \INPUT \textbf{I}, $a_{\text{initiator}}$
    \IF {approached by initiator}
        \IF [Swapping improves upon current choice]{$I_{a_n} \leq {I_{a_{\text{initiator}}}}$}
            \STATE response $\gets 1$
        \ELSE
            \STATE response $\gets 0$
        \ENDIF
    \ENDIF
    \OUTPUT response
    \NOTE $a_n$ is the current channel of user $n$
  \end{algorithmic}
\caption{The \textbf{respond} routine, performed by the user approached in a certain mini-frame}
\label{alg:resp}
\end{figure}

A user who is not the initiator and was not approached during the current mini-frame (for a detailed explanation see \secref{ssec:protocol}), uses the available time slot to transmit in her current channel and acquire reward accordingly, thus collecting a sample for her learning process. This step is carried out in \textbf{transmit\_and\_learn} (\figref{alg:trans-learn}).

\begin{figure}
  \begin{algorithmic}[1]
    \STATE $r\paren{t} \gets \textbf{transmit}\paren{a_n}$
    \STATE $r_{a_n} \gets r_{a_n} + r\paren{t}$
    \STATE $s_{a_n} \gets s_{a_n} + 1$
  \end{algorithmic}
\caption{The \textbf{transmit\_and\_learn} routine}
\label{alg:trans-learn}
\end{figure}

There are two special cases worth pointing out. First, we address the scenario that there is no initiator for the super frame (either zero users raised a flag, or more than one did so, see \figref{alg:choose-init}). Thanks to sensing, all users are aware of this, and can take advantage of the entire super-frame to execute the \textbf{transmit\_and\_learn} routine. While the users will not be able to improve their choice of channels during this super-frame, they will gather an increased number of learning samples and acquire more reward than they would during super-frames with an active initiator.

Our algorithm also addresses the scenario of swapping to an \emph{available} channel. In this case, the initiator need not transmit and wait for a response, and she simply switches to the channel once its turn on her preference list arrives.

\subsection{Coordination protocol}\label{ssec:protocol}
We now turn to the technical description of the coordination protocol that supports the CSM-MAB algorithm, based on the transmission and sensing capabilities of users described in \secref{ssec:limit-coord}.

\begin{figure}[!t]
\centering
\includegraphics[width=\columnwidth]{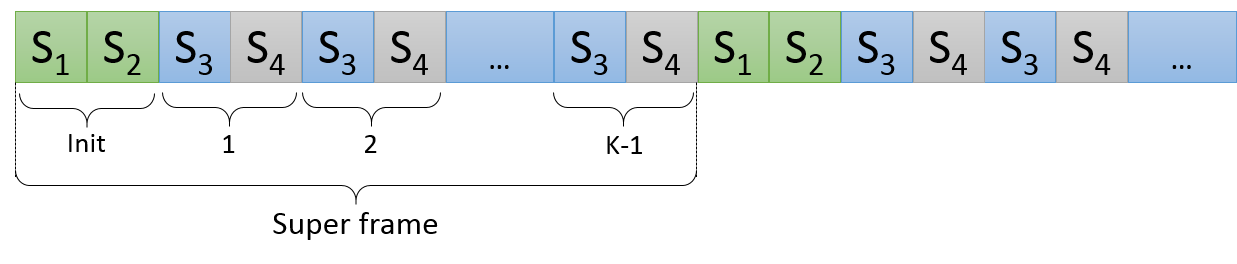}
\caption{Frame structure for CSM-MAB}
\label{fig:frames}
\end{figure}

In order to implement our protocol, we assume a synchronized, framed scheme. This assumption is fairly reasonable, and can be based on universal clocks such as Unix time. Time is divided into super-frames of $2K$ time slots each, with each super-frame consisting of an initialization mini-frame of two slots and $2\paren{K-1}$ coordination frames of two slots, as shown in \figref{fig:frames}.

\begin{figure}[!t]
\centering
\includegraphics[width=\columnwidth]{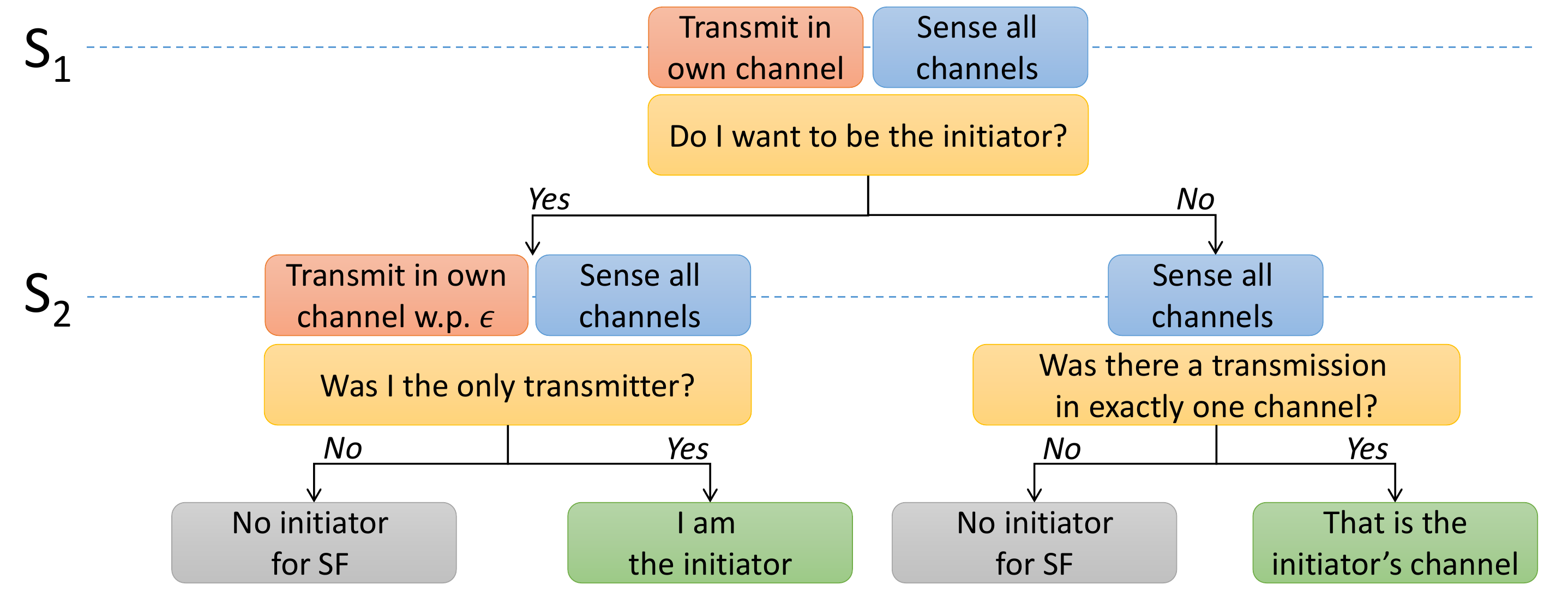}
\caption{Initialization mini-frame (\textbf{choose\_initiator} routine)}
\label{fig:alg1_init}
\end{figure}

During the initialization mini-frame, described in \figref{fig:alg1_init}, users transmit in their own channel and simultaneously sense all channels ($\text{S}_1$). This enables them to identify available channels, knowledge they will need if they become the initiator. Next, if they would like to become the initiator they transmit in their own channel and sense all channels. If they do not need to initiate a swap, they simply sense all channels ($\text{S}_2$). If there was a single transmitter in $\text{S}_2$, then she is the initiator for the super-frame and all users know her current channel, based on their sensing results. If no one transmitted during $\text{S}_2$ or more than one user transmitted, there is no initiator for the super-frame. Once again, users are aware of the situation by means of sensing, and they act accordingly (see \secref{ssec:alg-outline} for details).

Once the identity of the initiator has been determined, she manages the coordination process. In the worst case scenario, when sampling the channel that has a \emph{minimal} UCB index, a user will have $K-1$ channels on her preference list, to which she can upgrade. Therefore, the coordination process is made up of $K-1$ pairs of time slots, where each pair corresponds to a single entry on the initiator's preference list.

\begin{figure}[!t]
\centering
\includegraphics[width=\columnwidth]{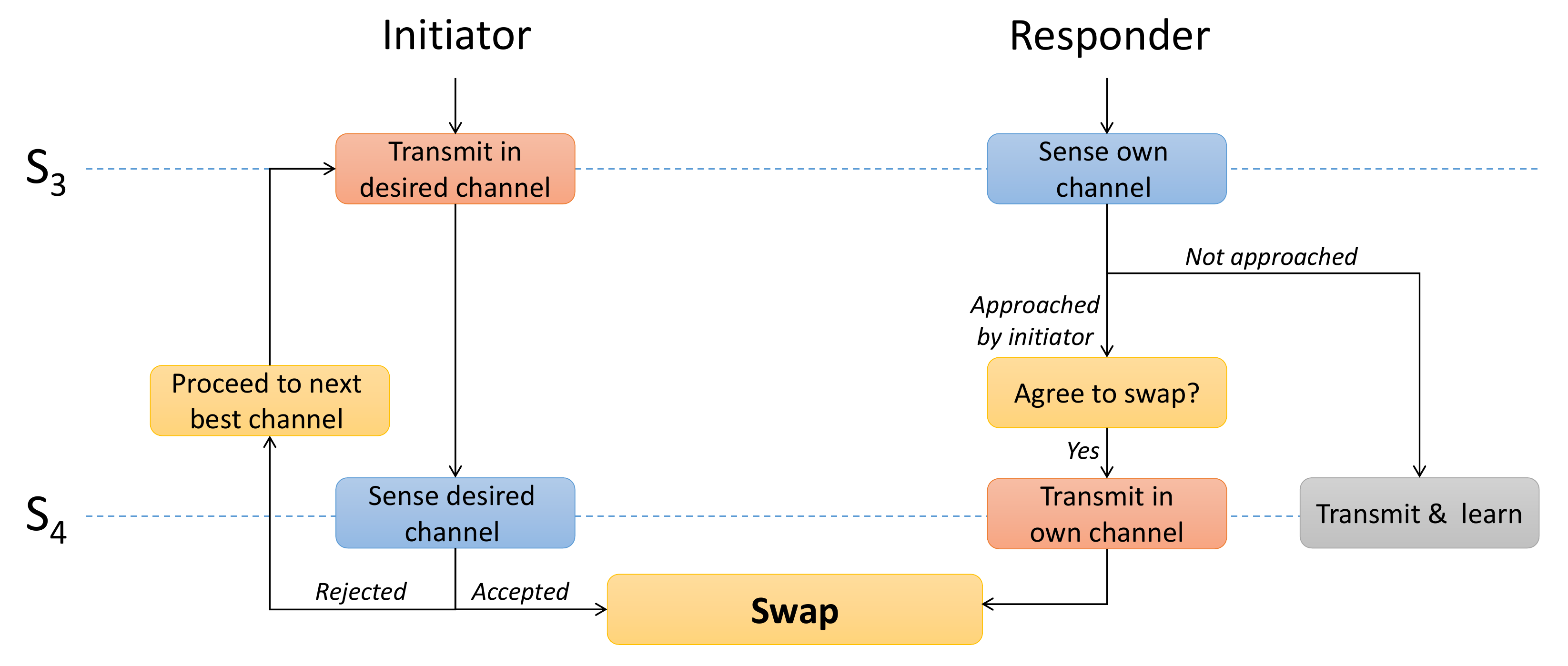}
\caption{Coordination process: initiator-responder dynamics (\textbf{coordinate\_swap} and \textbf{respond} routines)}
\label{fig:alg1_coord}
\end{figure}

During the coordination process, the initiator iterates over her preference list, starting with the index maximizing entry. For each entry, she transmits in the desired channel, while all other users sense for a signal in their respective channels ($\text{S}_3$). We refer to a user that senses a transmission in her channel as the responder. If the responder agrees to swap, she transmits in her own channel, which is simultaneously sensed by the initiator ($\text{S}_4$). A lack of transmission signals that the responder has declined, and prompts the initiator to proceed to the next entry on her list. Once the initiator succeeds in coordinating a swap or has probed all channels better than her current one, she will stop the coordination process. Once the super-frame ends, the process will re-start with a new initiator.

Time slots $\text{S}_4$ have a unique characteristic: they only require the initiator to perform sensing, and she only focuses her sensing on a single channel. Therefore, users who are not the initiator nor the responder may use this time slot to transmit in their current channel and collect reward, thus driving the learning process.

A technical issue worth addressing is that we assume transmission and sensing can be performed simultaneously. This assumption is reasonable in modern communication systems, in which users often have more than one antenna \cite{Jayaweera2014}.

To summarize this section, we would like to point out that the components of the CSM-MAB algorithm can be divided into two groups: components devoted to the learning process (\textbf{rank\_channels} and \textbf{transmit\_and\_learn}) and components contributing to the coordination of users in order to avoid collisions and ensure convergence to a stable configuration (\textbf{choose\_initiator}, \textbf{coordinate\_swap} and \textbf{respond}).

\section{Analysis of CSM-MAB}\label{sec:static-analysis}
We now turn to a theoretical analysis of the CSM-MAB algorithm. The result stated in \thmref{thm:static} shows that our algorithm meets the goals defined in \secref{ssec:goal}.

\begin{thm}\label{thm:static}
  Consider a system with $K$ channels and $N$ users, with channel rewards characterized by the matrix $\boldsymbol{\mu}$.
  Applying CSM-MAB (\figref{alg:CSM-MAB}) by all users will result in convergence to an orthogonal SMC:

  For all $\delta > 0$ there exists some finite $T\paren{\delta}$ such that for all time slots $t>T$, the probability of the system's being in an SMC is at least $1-\delta$, where
\begin{align*}
  T\paren{\delta} = O\paren{\log\paren{\frac{1}{\delta}},N,K^2}.
\end{align*}
\end{thm}

Achieving our goal consists of two aspects: orthogonality and stability. We address these aspects separately.
\subsection{Orthogonality}
The issue of orthogonality is rather simple: users need to reach a configuration in which there is at most one user sampling each channel. We ensure this by first applying the CFL algorithm of \cite{Leith2012}, in order to reach an orthogonal configuration (initially overlooking learning and stability). Once such a configuration has been reached, users only change their choice of channels in a coordinated manner, thus preserving orthogonality. We formalize this guarantee in the following proposition.

\begin{prop}
There exists a time $t_0>0$ so that the actions of users applying CSM-MAB are orthogonal (i.e., there is at most one user sampling each channel) for all $t>t_0$ with a probability of at least $1-\delta_0$.
\end{prop}

\begin{IEEEproof}
  Based on Theorem 1 of \cite{Leith2012}, the initial configuration reached after running the CFL algorithm is orthogonal with probability 1.
  The authors provide an upper bound on the distribution of stopping times, $\tau$:
  \begin{align*}
    \mP{\tau > k} = \alpha e^{-\gamma k},
  \end{align*}
  where $\alpha$ and $\gamma$ are some positive constants. The expected stopping time is therefore upper bounded by $\frac{\alpha e^{-\gamma}}{1-e^{-\gamma}}$.
  Setting $t_0 \teq \frac{2\alpha e^{-\gamma}}{1-e^{-\gamma}}$, the probability of not having reached an orthogonal configuration by time $t_0$ is at most $\delta_0 \teq e^{-2\frac{\alpha e^{-\gamma}}{1-e^{-\gamma}}}$.
  Once the system reaches an orthogonal configuration, users cannot switch channels without having coordinated the switch, as defined in \secref{ssec:alg-outline}.
\end{IEEEproof}

\subsection{Stability and potential}\label{ssec:potential}
In order to guarantee the second aspect, stability in the stable marriage sense, we define a system potential function. A single user's potential is the number of channels she would prefer over her current choice, had she known the true reward distributions. Formally, the potential of some user $n\in\set{1,\ldots,N}$ at time $t$ is defined as follows:
\begin{align}\label{eq:user_potential}
  \phi_n\paren{t} \triangleq \sum_{k = 1}^K \mathds{1}\set{\mu_{n,k} > \mu_{n,a_n\paren{t-1}}},
\end{align}
where $a_n\paren{t-1}$ is the action taken by user $n$ in the previous time step.

The system-wide potential is the sum of potentials over all users:
\begin{align}\label{eq:system_potential}
  \Phi\paren{t} \triangleq \sum_{n=1}^N \phi_n\paren{t}
\end{align}
An illustration of the potential appears in Tables 1 and 2.

\begin{table}[ht]
\centering
\caption{Table of users' channel rankings (first row represents best channel, last row represents worst). Cells highlighted in yellow and underline represent user's current choice.}
\label{tab:prefs}
\begin{tabular}{l|c|c|c|}
\cline{2-4}
                                 & \multicolumn{1}{l|}{\textbf{$U_1$}} & \multicolumn{1}{l|}{\textbf{$U_2$}} & \multicolumn{1}{l|}{\textbf{$U_3$}} \\ \hline
\multicolumn{1}{|l|}{\textbf{1}} & 1                                   & 2                                   & \cellcolor[HTML]{FFFE65}\underline{4}           \\ \hline
\multicolumn{1}{|l|}{\textbf{2}} & 2                                   & \cellcolor[HTML]{FFFE65}\underline{1}           & 1                                   \\ \hline
\multicolumn{1}{|l|}{\textbf{3}} & 4                                   & 3                                   & 2                                   \\ \hline
\multicolumn{1}{|l|}{\textbf{4}} & \cellcolor[HTML]{FFFE65}\underline{3}           & 4                                   & 3                                   \\ \hline
\end{tabular}
\end{table}

\begin{table}[ht]
\centering
\caption{User potentials corresponding to the configuration in \tabref{tab:prefs}.}
\label{tab:pot}
\begin{tabular}{|c|c|c|}
\hline
$\phi_1$ & $\phi_2$ & $\phi_3$ \\ \hline
3        & 1        & 0        \\ \hline
\end{tabular}
\end{table}

In terms of potential, a configuration is an SMC if no two users can swap channels and decrease their potential by doing so. We note that a stable configuration does not necessarily correspond to zero system-wide potential, since not all users might be able to achieve zero potential simultaneously, depending on network parameters. Also, a system may have several stable configurations, each characterized by a different potential.
Nevertheless, observing a system's potential does provide an indication regarding stability: once a system reaches a stable configuration, its potential will no longer change.

We would like to note that users' decisions are guided by UCB indices, while stability is examined with respect to true reward distributions. As a result, users do not always update their choice of channels in a way that matches the ground truth. This may lead to occasional increases in system potential, due to users' exploration or inaccurate statistics. In our proof we show that despite this, users ultimately converge to a stable configuration.

We use the potential function to prove convergence to an SMC, based on three observations:
\begin{enumerate}
  \item The maximal potential of a system with $N$ users and $K$ channels is $\Phi = N\paren{K-1}$.
  \item The potential $\Phi\paren{t}$ is monotonously non-increasing with high probability.
  \item As long as a system \emph{is not} in an SMC, changes in potential are bound to happen within finite time.
\end{enumerate}

The following lemmas formalize these observations, and are used as building blocks in the proof of \thmref{thm:static}. The proofs of the lemmas appear in \secref{app:static}.

\begin{lem}\label{lem:monoPot}
For all times $t$ for which $t > \frac{16K}{\Delta_{\min}^2}\ln t$, if a change in potential occurs, it is a decrease, with probability of at least $1-2t^{-4}$.
\end{lem}
Let $\Delta_{\min}$ be a distribution dependent constant.
In the appendix we derive the following bound for the minimal time for which the condition above holds:
\begin{align}\label{eq:t_min}
  t_{\min} = \paren{\frac{32K}{\Delta_{\min}^2}}^2.
\end{align}

\begin{lem}\label{lem:initiator}
If $\phi_n\paren{t} > 0$ for some user $n$, then her probability of becoming the next initiator is at least $\e\paren{1-\e}^{N-1}$.
\end{lem}

\begin{lem}\label{lem:finite_time_till_switch}
If the system is not in an SMC at some time $t>t_{\min}$, then the probability of a decrease in the potential occurring in the next super-frame is at least $1-\delta_1$, where
\begin{align*}
  \delta_1 = 1-\e\paren{1-\e}^{N-1} + 2t^{-4}.
\end{align*}
\end{lem}

We now turn to the proof of our main result.
\subsection{Proof of \thmref{thm:static}}
We model the convergence to an SMC using a Markov chain. Let $S_t$ denote the state of the system at time $t$:
\begin{align*}
  S_t =
  \begin{cases}
    1 & \text{if in SMC},\\
    0 & \text{else}.
  \end{cases}
\end{align*}
The probability of not being in an SMC, $T$ time slots after some point in time $t$, is therefore denoted by
\begin{align}\label{eq:delta}
  \mP{S_{t+T}=0}
  &= \cP{S_{t+T}=0}{S_t=0}\mP{S_t=0} \nonumber\\
  &\quad + \cP{S_{t+T}=0}{S_t=1}\mP{S_t=1}\nonumber\\
  &\leq \underbrace{\cP{S_{t+T}=0}{S_t=0}}_{P_1} + \underbrace{\cP{S_{t+T}=0}{S_t=1}}_{P_2}
\end{align}

Let us examine $P_1$ and $P_2$ separately.
\subsubsection{Bounding $P_2$}
The probability of the event of moving from an SMC to an unstable configuration is bounded by the probability of a single increase in potential in the time interval $\sbrk{t,t+T}$:
\begin{align*}
  P_2 \leq \mP{\Phi_{t+T} = \Phi_{t}+1} \leq 2t^{-4},
\end{align*}
based on \lemref{lem:monoPot}.
\subsubsection{Bounding $P_1$}
\begin{align*}
  P_1 = 1 - \cP{S_{t+T}=1}{S_t = 0}
\end{align*}
A transition from an unstable configuration to an SMC can occur if the difference between potential decreases and increases is anywhere in the range $\sbrk{1,\ldots,N\paren{K-1}}$, depending on the potential at time $t$ and on the potential of the SMC reached. Therefore, we can bound the probability $\cP{S_{t+T}=1}{S_t = 0}$ by the worst case, where the difference is maximal: $N\paren{K-1}$.

Denoting the number of increases in potential in the time interval $\sbrk{t_1,t_2}$ by $I_{\sbrk{t_1,t_2}}$,
\begin{align}\label{eq:proofThm1_P1comp}
  \mathbb{P}&\left[\given{S_{t+T}=1}{S_t = 0}\right]\nonumber\\
  &\geq \mP{D_{\sbrk{t,t+T}} - I_{\sbrk{t,t+T}} > N\paren{K-1}} \nonumber\\
  &= \mP{D_{\sbrk{t,t+T}} > N\paren{K-1} + I_{\sbrk{t,t+T}}} \nonumber\\
  &= \sum_{i=0}^{\infty}\cP{D_{\sbrk{t,t+T}} > N\paren{K-1} + i}{I_{\sbrk{t,t+T}} = i}\mP{I_{\sbrk{t,t+T}} = i} \nonumber\\
  &\geq \mP{D_{\sbrk{t,t+T}} > N\paren{K-1} + 0}\mP{I_{\sbrk{t,t+T}} = 0}.
\end{align}

We now bound the probability of zero potential increases from below. In order to do so, we will need a bound on the probability distribution of potential increases, $\mP{I_{\sbrk{t,t+T}} = i}$. A single increase in potential occurs if two conditions hold:
\begin{itemize}
  \item An initiator emerges
  \item The statistics for \emph{both} users are incorrect
\end{itemize}

\begin{figure*}[!t]
\normalsize
\setcounter{MYtempeqncnt}{\value{equation}}
\setcounter{equation}{6}
\begin{align}\label{eq:T_delta_1}
  \frac{T_{SF}}{T}\paren{\frac{T}{T_{SF}}\paren{\e\paren{1-\e}^{N-1} - 2t_{\min}^{-4}} - N\paren{K-1}}^2
  &= \half\log \paren{\frac{1}{\delta - 6t_{\min}^{-4}}} \nonumber\\
  \paren{\sqrt{\frac{T_{SF}}{T}}\frac{T}{T_{SF}}\paren{\e\paren{1-\e}^{N-1} - 2t_{\min}^{-4}} - \sqrt{\frac{T_{SF}}{T}}N\paren{K-1}}^2
  &= \half\log \paren{\frac{1}{\delta - 6t_{\min}^{-4}}} \nonumber\\
  \sqrt{\frac{T}{T_{SF}}}\paren{\e\paren{1-\e}^{N-1} - 2t_{\min}^{-4}}
  - \sqrt{\frac{T_{SF}}{T}}N\paren{K-1}
  &= \paren{\half\log \paren{\frac{1}{\delta - 6t_{\min}^{-4}}}}^\half.
\end{align}
\setcounter{equation}{\value{MYtempeqncnt}}
\hrulefill
\vspace*{4pt}
\end{figure*}

Deriving an upper bound on the probability of an initiator emerging, denoted by $P_{\text{init}}$:
\begin{align*}
  P_{\text{init}} &\leq \sum_{n=1}^N \e\paren{1-\e}^{n-1} \\
  &= \e\sum_{n=0}^{N-1}\paren{1-\e}^{n} \\
  &= \e \frac{1-\paren{1-\e}^N}{1-\paren{1-\e}} \\
  &= 1 - \paren{1-\e}^N.
\end{align*}
Based on the derivation leading to \lemref{lem:monoPot}, the probability that both users' statistics are wrong is upper bounded by $2t^{-4}$. Therefore, the probability of a single increase in potential is bounded by $2t^{-4}\paren{1 - \paren{1-\e}^N}$, and the probability of $i$ increases is no more than $P_i\triangleq\paren{2t^{-4}}^i\paren{1 - \paren{1-\e}^N}^i$.
We can now bound the probability of encountering potential increases:
\begin{align*}
  \mP{I_{\sbrk{t,t+T}} = 0}
  &= 1 - \mP{I_{\sbrk{t,t+T}} \geq 0} \\
  &= 1 - \sum_{i=1}^{\infty}\mP{I_{\sbrk{t,t+T}} = i}\\
  &\geq 1 - \sum_{i=1}^{\infty} P_i \\
  &\geq 1 - \frac{2t^{-4}\paren{1 - \paren{1-\e}^N}}{1-2t^{-4}\paren{1 - \paren{1-\e}^N}} \\
  &\geq 1 - 4t^{-4}\paren{1 - \paren{1-\e}^N}.
\end{align*}
In order to simplify the expression, and since the dependency on $\e$ here is rather weak, in the sequel we will use the bound
\begin{align}\label{eq:proofThm1_zeroIncs}
  \mP{I_{\sbrk{t,t+T}} = 0} \geq 1 - 4t^{-4}.
\end{align}

We now turn back to \eqref{eq:proofThm1_P1comp}, and continue by bounding the potential decrease term from below. Let us consider a Binomial random process that lower bounds the process of potential decreases. The success probability is (a lower bound on) the probability of a potential decrease in a single super-frame, and the number of experiments is the number of whole super-frames in the interval $\sbrk{t,t+T}$. Applying Hoeffding's inequality to a Binomial random variable $X$ with success probability $p$ over $n$ experiments yields an upper bound on the probability of acquiring exactly or less than $k$ successes:
\begin{align*}
  \mP{X\leq k} \leq e^{-2\frac{\paren{np-k}^2}{n}}.
\end{align*}
In our case, we use \lemref{lem:finite_time_till_switch} to determine that $p\triangleq \e\paren{1-\e}^{N-1} - 2t^{-4}$, the number of experiments is $n = \floor{\frac{T}{T_{SF}}}$ and $k = N\paren{K-1}$.
Therefore,
\begin{align*}
\mathbb{P}&\left\{D_{\sbrk{t,t+T}} > N\paren{K-1}\right\}\\
  &= 1 - \mP{D_{\sbrk{t,t+T}} \leq  N\paren{K-1}} \\
  &\geq 1 - e^{-2\frac{\paren{\frac{T}{T_{SF}}\paren{\e\paren{1-\e}^{N-1} - 2t^{-4}} - N\paren{K-1}}^2}{\paren{\frac{T}{T_{SF}}}}},
\end{align*}
where we dropped the ``floor'' operator for the sake of clarity.

Returning to \eqref{eq:proofThm1_P1comp}:
\begin{align*}
  \cP{S_{t+T}=1}{S_t = 0} &\geq \paren{1-e^{-2\alpha}}\paren{1 - 4t^{-4}} \\
  &= 1 - e^{-2\alpha} + 4t^{-4}e^{-2\alpha} - 4t^{-4} \\
  &\geq 1 - e^{-2\alpha} - 4t^{-4},
\end{align*}
where $\alpha = \frac{\paren{\frac{T}{T_{SF}}\paren{\e\paren{1-\e}^{N-1} - 2t^{-4}} - N\paren{K-1}}^2}{\paren{\frac{T}{T_{SF}}}}$.
\subsubsection{Combining the bounds}
Backing up all the way to $P_1$ and $P_2$ of \eqref{eq:delta}, we have that
\begin{align*}
  \mP{S_{t+T}=0} \leq e^{-2\alpha} + 6t^{-4}.
\end{align*}
We can now denote $\delta = e^{-2\alpha} + 6t^{-4}$.

\subsubsection{Deriving the explicit dependence of $T$ on $\delta$}
Since our convergence guarantee holds for $t>t_{\min}$, we can substitute $t_{\min}$ wherever $t$ appears:
\begin{align*}
  \delta    & = e^{-2\alpha} + 6t_{\min}^{-4}\\
  \alpha    & = \half\log \paren{\frac{1}{\delta - 6t_{\min}^{-4}}}
\end{align*}
Substituting the value of $\alpha$ and rearranging, we have \eqref{eq:T_delta_1}.
\addtocounter{equation}{1}
Let us observe a simplified form of \eqref{eq:T_delta_1}:
\begin{align}\label{eq:simpT}
  a\sqrt{T} - \frac{b}{\sqrt{T}} = c,
\end{align}
where
\begin{align*}
  a &= \frac{1}{\sqrt{T_{SF}}}\paren{{\e\paren{1-\e}^{N-1} - 2t_{\min}^{-4}}}\\
  b &= \sqrt{T_{SF}}N\paren{K-1}\\
  c &= \paren{\half\log \paren{\frac{1}{\delta - 6t_{\min}^{-4}}}}^\half.
\end{align*}
\begin{figure*}[!t]
\normalsize
\setcounter{MYtempeqncnt}{\value{equation}}
\setcounter{equation}{8}
\begin{align}\label{eq:T_delta_full}
  T\paren{\delta} = t_{\min} + \frac{T_{SF}}{{\e\paren{1-\e}^{N-1} - 2t_{\min}^{-4}}}
  \sbrk{\frac{1}{4\paren{{\e\paren{1-\e}^{N-1} - 2t_{\min}^{-4}}}}\log \paren{\frac{1}{\delta - 6t_{\min}^{-4}}}+2N\paren{K-1}}.
\end{align}
\setcounter{equation}{\value{MYtempeqncnt}}
\addtocounter{equation}{1}
\hrulefill
\vspace*{4pt}
\end{figure*}
Substituting $Q = \sqrt{T}$ into \eqref{eq:simpT}:
\begin{align*}
  aQ^2 - cQ - b &= 0 \\
  Q_{1,2} = \frac{c \pm\sqrt{c^2 + 4ab}}{2a}\\
  Q = \frac{c +\sqrt{c^2 + 4ab}}{2a},
\end{align*}
where we took the positive solution because of the way $Q$ is defined.
Back to the equation in terms of $T$:
\begin{align*}
  T &= \paren{\frac{c +\sqrt{c^2 + 4ab}}{2a}}^2 \\
  &= \frac{c^2}{4a^2} + \frac{c^2+4ab}{4a^2} + \frac{c\sqrt{c^2+4ab}}{2a^2} \\
  &= \frac{c^2}{2a^2} + \frac{b}{a} + \frac{c\sqrt{c^2+4ab}}{2a^2} \\
  &\leq \frac{c^2}{2a^2} + \frac{b}{a} + \frac{c}{2a^2}\sqrt{\paren{c+\frac{2ab}{c}}^2}\\
  &= \frac{c^2}{2a^2} + \frac{b}{a} + \frac{c}{2a^2}\paren{c+\frac{2ab}{c}}\\
  &= \frac{c^2}{a^2} + \frac{2b}{a}.
\end{align*}
Plugging in the values of $a,b$ and $c$ and summarizing our derivation, we have that for all $\delta>0$, the system is guaranteed to reach an SMC with a probability of at least $1-\delta$ within $T\paren{\delta}$ time slots. The full expression for $T\paren{\delta}$ appears in \eqref{eq:T_delta_full}.
Roughly,
\begin{align*}
  T\paren{\delta} = O\paren{\log\paren{\frac{1}{\delta}},N,K^2},
\end{align*}
where the quadratic dependence on $K$ stems from the fact that $T_{SF} = 2K$. The parameter $\delta$ reflects a (small) failure probability, as customary in PAC learning settings.

\section{D-CSM-MAB algorithm}\label{sec:D-CSM-MAB}
In this section we introduce a variant of the CSM-MAB algorithm suited for systems with a dynamic number of users.
Such a scenario is most common in communication networks of independent users, and our architecture handles it with minor additional effort.
Departure of users can be handled without any changes to the original algorithm, while their arrival requires some adjustments, as described below.

In designing our algorithm we make a single assumption, that is required for arriving users to join the system smoothly.
\begin{assum}
  During a single super-frame, the number of arriving users is no more than 1.
\end{assum}
As shown below, the length of a single super-frame is $2K + 1$. In practical systems this is a very short time interval, typically less than a millisecond.

\subsection{Algorithm outline}
\begin{figure}[!t]
\centering
\includegraphics[width=\columnwidth]{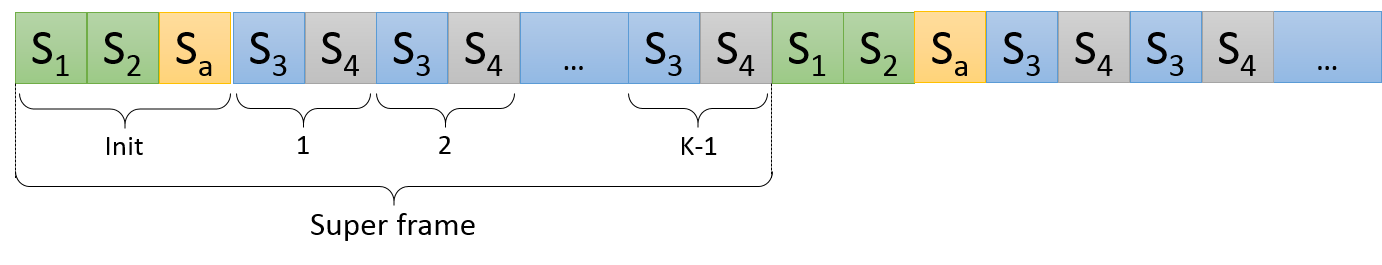}
\caption{Frame structure for D-CSM-MAB, new slot (S$_\text{a}$) in yellow}
\label{fig:frames-d}
\end{figure}

Supporting a dynamic number of users entails a small change in the frame structure, shown in \figref{fig:frames-d}: a single slot per super-frame is added, as part of the initialization mini-frame.

The D-CSM-MAB algorithm is presented in \figref{alg:D-CSM-MAB}, with the modification highlighted in blue on \lineref{line:newbie}.
\begin{figure}
  \begin{algorithmic}[1]
    \STATE \textbf{apply\_CFL}
    \FORALL{time slots $t$}
        \IF {Beginning of super-frame}
            \STATE \textbf{rank\_channels}
            \STATE \textcolor{blue}{\textbf{newbie\_joins}} \label{line:newbie}
            \STATE \textbf{choose\_initiator}
        \ELSE
            \IF {$n$ is the initiator}
                \STATE \textbf{coordinate\_swap}
            \ELSIF {$n$ was approached by initiator}
                \STATE \textbf{respond}
            \ELSE
                \STATE \textbf{transmit\_and\_learn}
            \ENDIF
        \ENDIF
    \ENDFOR
  \end{algorithmic}
\caption{The D-CSM-MAB algorithm for user $n$}
\label{alg:D-CSM-MAB}
\end{figure}
The \textbf{newbie\_joins} routine is executed only by the arriving user, and is described in \figref{alg:newbie}.
\begin{figure}
  \begin{algorithmic}[1]
    \INPUT available
    \IF [an available channels exists]{available $\neq \phi$}
        \STATE channel $\gets$ \textbf{rand}(Uniform,available)
    \ENDIF
  \end{algorithmic}
\caption{The \textbf{newbie\_joins} routine}
\label{alg:newbie}
\end{figure}

\subsection{Detailed flow}
An arriving user faces the challenge of finding an initial channel to transmit in, without colliding with existing users. She does so by examining the sensing result of $\text{S}_1$ (see \secref{ssec:protocol}), thus determining which channels are available. If all channels are taken, she will wait for an entire super-frame before repeating the process, in the hope that one of the users will have left by then.

If there is at least one available channel, the arriving user will choose one uniformly at random. She will then signal her choice by transmitting in that channel during time slot $\text{S}_\text{a}$. The rest of the users will sense the spectrum during this time slot, thus becoming aware of the new user's arrival and the occupation of an additional channel.

Starting with the following super-frame, the user will apply the D-CSM-MAB algorithm as a ``veteran'', employing the initiator-responder mechanism in order to reach the best possible channel. 

\section{D-CSM-MAB analysis}\label{sec:dynamic-analysis}
Our analysis of the D-CSM-MAB algorithm considers arrivals and departures separately, as they have quite an opposite effect on the system.

\subsection{Analysis of user arrival}
Users' arrivals result in an unstable configuration, since they are interested in swapping channels until their learning process converges. Our analysis quantifies this phenomenon.
\begin{thm}
  Let $S$ be a system with $K$ channels and $N$ users. If $S$ is in an SMC at some time $t$ and a new user joins the system, the new system $S'$ will settle into an SMC within no more than $T\paren{K,N,\delta}$ time slots with a probability of at least $1-\delta$.
\end{thm}
The basic idea behind our result is that the arriving user will join the system by occupying one of the $K-N$ available channels. She will then attempt to learn the statistics of all the channels, resulting in her ``hopping'' between the $K-N$ channels that were available upon her arrival. Eventually, she will converge to one of them and the system will be in an SMC once again.

\begin{IEEEproof}
We would like to bound the probability that the new system, $S'$ has not settled into an SMC by time $T$.
The system's return to an SMC is determined by the arriving user: she needs to gather a sufficient number of samples from each of the available $K-N$ channels, before she can prefer one of them. Only then can we consider her state a stable marriage, and the entire system's configuration an SMC.

Let us denote the series of points in time in which the arriving user is chosen as initiator by $\set{t_i},i=1,\ldots,I$, where $L$ is the size of this set.
The intervals between consecutive $t_i$'s are denoted by $\Delta_i = t_{i}-t_{i-1},i = 1..L$, where $t_0=t$.

\subsubsection{Proof outline}
The arriving user might \emph{not} reach a stable state by time $T$, for for two reasons: either she was not chosen as initiator often enough, i.e. $L<K-N$, or if she was chosen often enough, the intervals $\Delta_i$ were not distributed well. If intervals are too long, they may prohibit a fair division of samples between the arms that need to be learned. For an illustration of this concept see \figref{fig:dynamic-proof}. The top part of the figure displays an unbalanced interval distribution: some of the intervals are very short ($\beta_1$), and one of the intervals is very long ($\beta_2$). If $\beta_1$ is small with respect to the minimal number of samples required, then some of the $K-N$ arms may not be sampled sufficiently, resulting in an unstable state at time $T$. A more favourable distribution appears in the bottom figure, assuming that $K-N \leq 5$. In short, we would like to ensure that interval lengths are not too long with high probability.

\begin{figure}[!t]
\centering
\includegraphics[width=0.8\columnwidth]{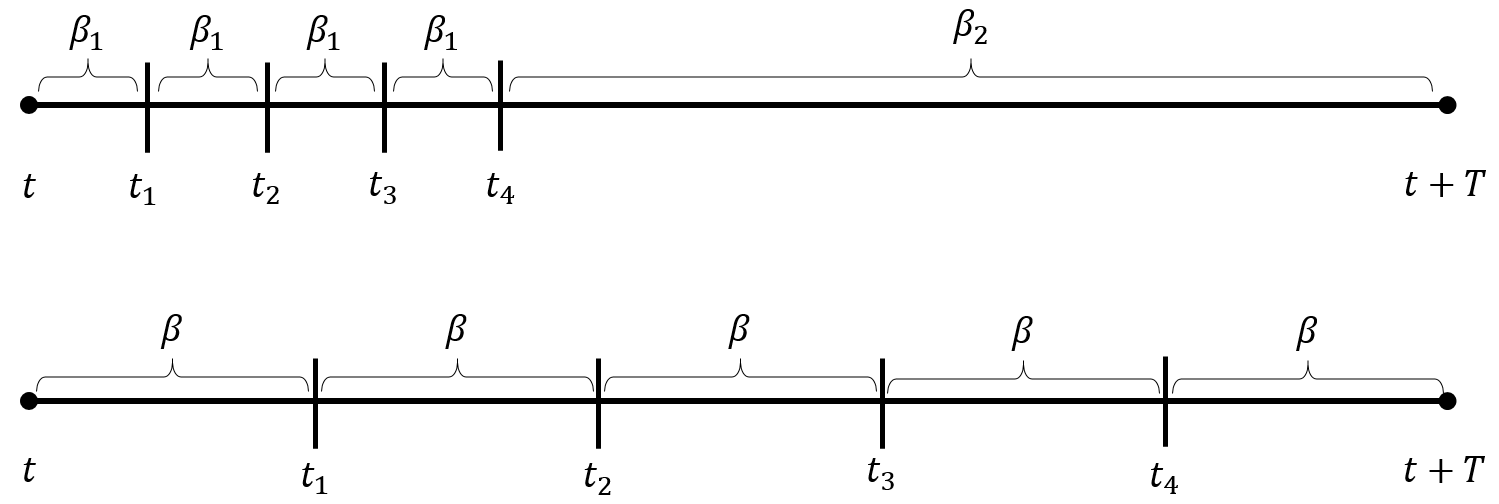}
\caption{Possible distributions of $\Delta_i$ intervals}
\label{fig:dynamic-proof}
\end{figure}

We note that since the system is assumed to have been in an SMC prior to time $t$, the other users will not be interested in swapping channels with the arriving user, and so she needs to be the initiator in order to move between channels.

\subsubsection{Required samples}
Based on the derivation in \secref{app:static-lem1}, the minimal number of samples per channel required is as stated in \eqref{eq:s_min}:
\begin{align}\label{eq:s_min-dyn}
  s_{\min} \triangleq \frac{8\ln T}{\Delta_{\min}^2},
\end{align}
where $\Delta_{\min}$ is a constant that depends on the channels' reward distributions (and is not related to the $\Delta_i$'s defined above).

\subsubsection{Intervals}
Let us assume the worst case, in which all intervals are of equal length, $\beta$, where length is measured in super-frames.
For the sake of clarity, we omit ceiling and floor operators, and assume all fraction results are rounded.
The number of intervals of length $\beta$ in the period $\sbrk{t,t+T}$ is $L = \frac{T}{\beta}$.
Short intervals do not interfere with the learning process, so the size of $\beta$ should only be bounded from above.
The numbers of intervals dedicated to a single channel can differ by no more than one, and so it is either $\frac{T}{\beta\paren{K-N}}$ or $\frac{T}{\beta\paren{K-N}}+1$.
The minimal number of intervals is therefore $\frac{T}{\beta\paren{K-N}}$, and since each super-frame contributes $2K-2$ samples, the condition for acquiring enough samples is
\begin{align*}
  \frac{T}{\beta\paren{K-N}}\paren{2K-2} \geq s_{\min},
\end{align*}
or, re-written:
\begin{align}\label{eq:dynamic-beta}
  \beta \leq \frac{T}{s_{\min}}\frac{2K-2}{K-N}\triangleq \beta_{\max}.
\end{align}
The distribution of interval lengths is geometric, with a bounded success probability $p$: $\e \leq p \leq \e\paren{1-\e}^N$.
Based on the cumulative density function of geometric random variables, we have that for some $k$
\begin{align*}
  \mP{\Delta_i \leq k} &= 1 - (1-p)^{K+1} \\
  \mP{\Delta_i > k} &= (1-p)^{K+1} \leq \paren{1-\e}^{K+1}.
\end{align*}
Plugging in the condition of \eqref{eq:dynamic-beta}, we have that
\begin{align*}
  \mP{\Delta_i > \beta_{\max}}
  &= \mP{\Delta_i > \frac{T}{s_{\min}}\frac{2K-2}{K-N}} \\
  &\leq \paren{1-\e}^{\frac{T}{s_{\min}}\frac{2K-2}{K-N}+1}.
\end{align*}

\begin{figure*}[!t]
\normalsize
\setcounter{MYtempeqncnt}{\value{equation}}
\setcounter{equation}{12}
\begin{align}\label{eq:T_delta_dyn}
  T\paren{\delta} = t_{\min} + \frac{T_{SF}}{{\e\paren{1-\e}^{N-1} - 2t_{\min}^{-4}}}
  \sbrk{\frac{1}{4\paren{{\e\paren{1-\e}^{N-1} - 2t_{\min}^{-4}}}}\log \paren{\frac{1}{\delta - 6t_{\min}^{-4}}}+N\paren{N-1}}.
\end{align}
\setcounter{equation}{\value{MYtempeqncnt}}
\hrulefill
\vspace*{4pt}
\end{figure*}

\subsubsection{Putting it all together}
We can now derive the relationship between the error probability, $\delta$, and the length of the interval, $T$.
\begin{align*}
  \delta &= \frac{1}{1-\e}^{-\frac{T}{s_{\min}}\frac{2K-2}{K-N}+1} \\
  \ln\frac{1}{\delta} &= \paren{\frac{T}{s_{\min}}\frac{2K-2}{K-N}+1}\ln\frac{1}{1-\e} \\
  T &= s_{\min}\frac{K-N}{2K-2}\paren{\frac{\ln\frac{1}{\delta}}{\ln\frac{1}{1-\e}}-1}.
\end{align*}
Plugging in the value of $s_{\min}$ from \eqref{eq:s_min-dyn}:
\begin{align*}
  \frac{T}{\ln T} = \frac{4}{\Delta_{\min}^2}\frac{K-N}{K-1}\paren{\frac{\ln\frac{1}{\delta}}{\ln\frac{1}{1-\e}}-1}
\end{align*}
Based on equation (5.2.4) of \cite{Hardy1952}, we have that $\ln T < 2\sqrt{T}$, and therefore a looser condition would be
\begin{align}
  T \geq \paren{\frac{4}{\Delta_{\min}^2}\frac{K-N}{K-1}\paren{\frac{\ln\frac{1}{\delta}}{\ln\frac{1}{1-\e}}-1}}^2.
\end{align}
\end{IEEEproof}

\subsection{Analysis of user departure}
Our analysis of convergence to an SMC after a user's departure focuses on the worst case scenario and follows the proof of \thmref{thm:static}.
\begin{thm}
  Let $S$ be a system with $K$ channels and $N$ users. If $S$ is in an SMC at some time $t$ and one of the users leaves the system, the new system $S_1$ will settle into an SMC
within no more than $T\paren{N,\delta}$ time slots with a probability of at least $1-\delta$.
\end{thm}
The convergence of the system to a new SMC depends heavily on its potential before the departure. The higher the system potential, the longer it will take the system to converge.
Our proof therefore involves an upper bound on the potential of a system in an SMC.
\begin{lem}\label{lem:max_pot}
  Let $S$ be a system with $K$ channels and $N$ users. If $S$ is in an SMC, then its system wide potential is no more than $\Phi_{\max}\triangleq \half N\paren{N-1}$.
\end{lem}

The example in \tabref{tab:prefs-dyn} and \tabref{tab:pot-dyn} illustrates the claim of \lemref{lem:max_pot}. The most difficult scenario for our problem occurs when users' preferences are all the same, as shown in the example. In such a scenario, user dissatisfaction is maximal, and the potential remains high even for stable configurations.

\begin{table}[ht]
\centering
\caption{Table of users' channel rankings (first row represents best channel, last row represents worst). Cells highlighted in yellow and underline represent user's current choice.}
\label{tab:prefs-dyn}
\begin{tabular}{l|c|c|c|c|}
\cline{2-5}
                                 & \multicolumn{1}{l|}{\textbf{$U_1$}} & \multicolumn{1}{l|}{\textbf{$U_2$}} & \multicolumn{1}{l|}{\textbf{$U_3$}} & \multicolumn{1}{l|}{\textbf{$U_4$}}\\ \hline
\multicolumn{1}{|l|}{\textbf{1}} & \cellcolor[HTML]{FFFE65}\underline{1}                                   & 1                                   & 1   & 1           \\ \hline
\multicolumn{1}{|l|}{\textbf{2}} & 2                                   & \cellcolor[HTML]{FFFE65}\underline{2}  & 2         & 2                                   \\ \hline
\multicolumn{1}{|l|}{\textbf{3}} & 3                                   & 3                                   & \cellcolor[HTML]{FFFE65}\underline{3} & 3                                   \\ \hline
\multicolumn{1}{|l|}{\textbf{4}} & 4           & 4                         & 4          & \cellcolor[HTML]{FFFE65}\underline{4}                                   \\ \hline
\end{tabular}
\end{table}

\begin{table}[ht]
\centering
\caption{User potentials corresponding to the configuration in \tabref{tab:prefs-dyn}.}
\label{tab:pot-dyn}
\begin{tabular}{|c|c|c|c|}
\hline
$\phi_1$ & $\phi_2$ & $\phi_3$ & $\phi_4$\\ \hline
0        & 1        & 2   & 3      \\ \hline
\end{tabular}
\end{table}

\begin{IEEEproof}
  The worst case scenario in terms of potential of a stable configuration occurs when users' rankings are equal.
  Formally, let $\ell_{n,k}$ be the rank of channel $k$ for user $n$, meaning that, for example, $k_n^* = \argmax_k\ell_{n,k}$ is the best channel for user $n$.
  The vector of all rankings of user $n$ is $\ell_n = \paren{\ell_{n,1},\ldots,\ell_{n,K}}$, and users' rankings are equal when $\ell_n  = \ell_m, \forall n,m\in\set{1,\ldots,N}$.

  The potential of any stable configuration in this scenario is
  \begin{align*}
    \Phi_{\max} = 0 + 1 + \ldots + N-1 = \sum_{i = 0}^{N-1} \paren{N-i} = \half N\paren{N-1}.
  \end{align*}
\end{IEEEproof}

Plugging the upper bound on the potential into the proof of \thmref{thm:static} yields a bound on the convergence time to a new SMC, presented in \eqref{eq:T_delta_dyn}.
\addtocounter{equation}{1}

\section{Experiments}\label{sec:experiments}
In order to demonstrate the performance of our algorithms, we present simulation results of a multi-user communication network.
The network consists of $N$ users and $K$ channels, where $N$ and $K$ are simulation parameters.
The users follow the algorithm introduced in \secref{sec:CSM-MAB} and cannot communicate directly in any way.
Upon transmitting in a certain channel, users observe a signal that reflects the quality of their transmission.
In practical terms this can be the channel throughput, an ``ACK'' binary signal or any other performance criterion.
The observed signal, modelled by the reward, drives the learning process.

As explained in \secref{ssec:system}, the reward acquired in a transmission depends both on the transmitting user's identity and on the channel index.
The reward a transmission yields is therefore drawn from the reward distribution for the appropriate (user,channel) pair.
In our experiments we chose to model ``ACK'' signals acquired by users.
These signals are widespread in pairwise communications, and are used by the receiver to notify the transmitter of a successful transmission.
The reward distributions we use are therefore binary, with a parameter (expected value) $\mu_{n,k}$.
The values of the expected rewards are drawn uniformly from the interval $\sbrk{0,1}$ at the beginning of an experiment.

In this section we present several results, examining different aspects of system performance.
First, we address the number of times users change their policy, namely switch channels.
A low rate of policy changes attests to the stability of a configuration, and we would therefore like such changes to become scarce over time.
From a practical point of view, modern communication involves transmission of long, ``heavy'' multi-media data streams.
Frequent interruptions are detrimental to such communications, and we therefore prefer long stretches of time between channel switches.

\begin{figure}[!t]
\centering
\includegraphics[width=\columnwidth]{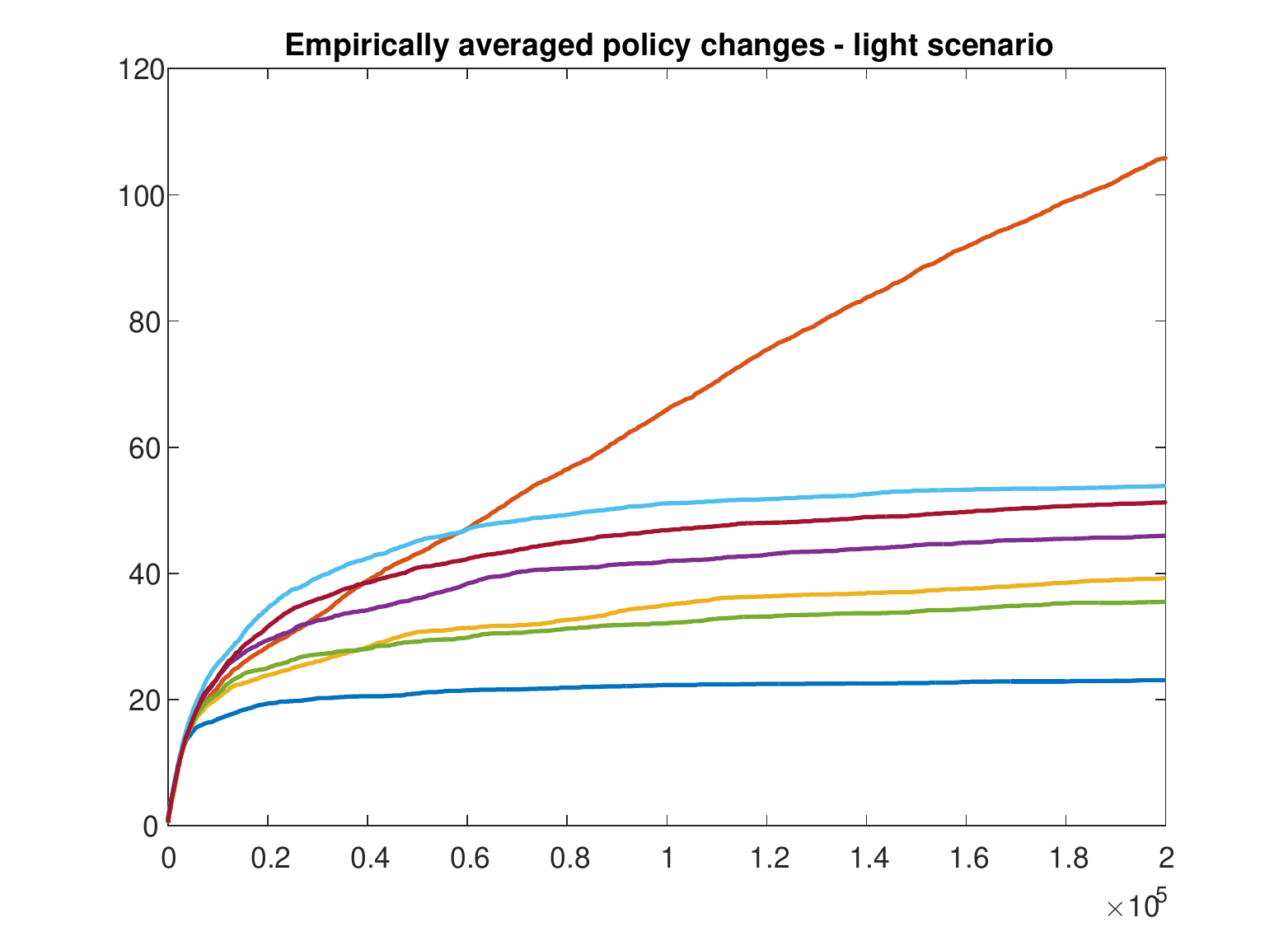}
\caption{Policy switches - light scenario: K=10, N=7}
\label{fig:PolicySwitch1}
\end{figure}

\begin{figure}[!t]
\centering
\includegraphics[width=\columnwidth]{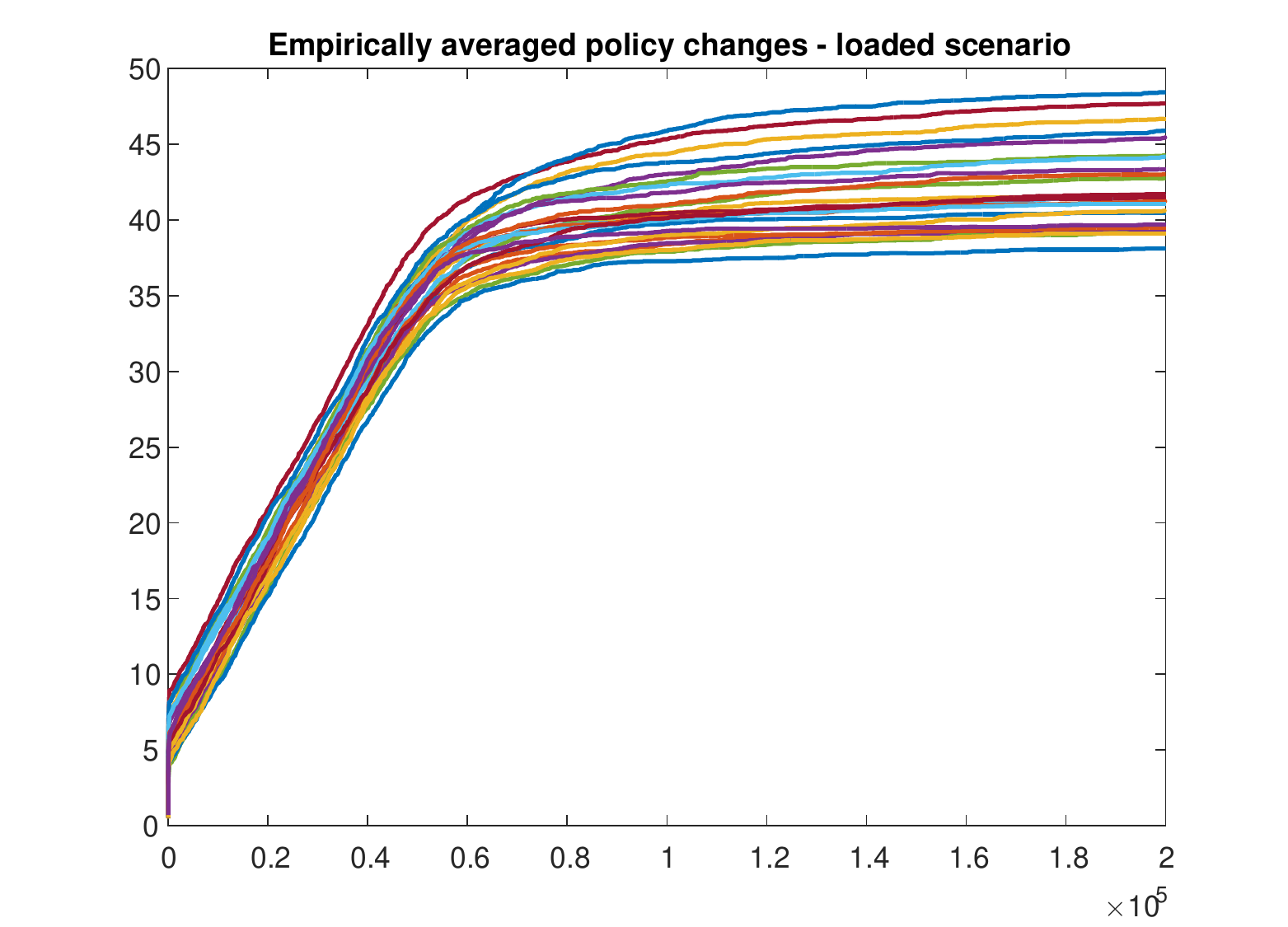}
\caption{Policy switches - loaded scenario: K=25, N=25}
\label{fig:PolicySwitch2}
\end{figure}

\figref{fig:PolicySwitch1} and \figref{fig:PolicySwitch2} display the cumulative number of channel switches per user over the course of an entire experiment, that lasts $T=200,000$ time slots. The results are averaged over 50 repetitions of the same setup. The light scenario is composed of 7 users accessing 10 channels, while the loaded one includes 25 users and 25 channels. Several interesting phenomena can be observed in these graphs. First, convergence, in the form of few channel switches, is achieved in both scenarios, as is evident from the plateau in the graphs. In addition, convergence for the loaded scenario is delayed when compared to the light one. This is due to two factors: the increased number of channels that lengthens the learning period, and the fact that the number of users is equal to the number of channels, increasing the complexity of switching channels. Finally, \figref{fig:PolicySwitch1} displays a interesting situation, in which one of the users continues to switch channels long after her fellow users have converged to a single channel. This is a result of this user's reward distribution: there are two channels whose expected rewards for this user are very close. To be more specific, user 2 has an expected reward of 0.9626 in channel 7, and and expected reward of 0.9569 in channel 2. The difference between these values drives the user's learning process; since they are very close ($\delta = 0.0057$), the user will need many samples to choose between the channels. This is a result of the definition of the UCB index, used in \algref{alg:rank-channels}. In terms of the acquired reward, these switches do not hurt performance, since the channels are very close. However, this situation may be undesirable in terms of stability, and we propose a solution for it in \secref{sec:conclusion}.

Next, we present the potential of the system over time, as defined in \secref{sec:static-analysis}, equations \eqref{eq:user_potential} and \eqref{eq:system_potential}. This measure also serves as evidence of the system's convergence to a stable configuration. As before, we examine scenarios with light and heavy loads. The potential exhibits the decay we expected, along with the previously observed difference between the two scenarios. The shaded area around the plots represents the variance over experiment repetitions, and is rather small.

\begin{figure}[!t]
\centering
\includegraphics[width=\columnwidth]{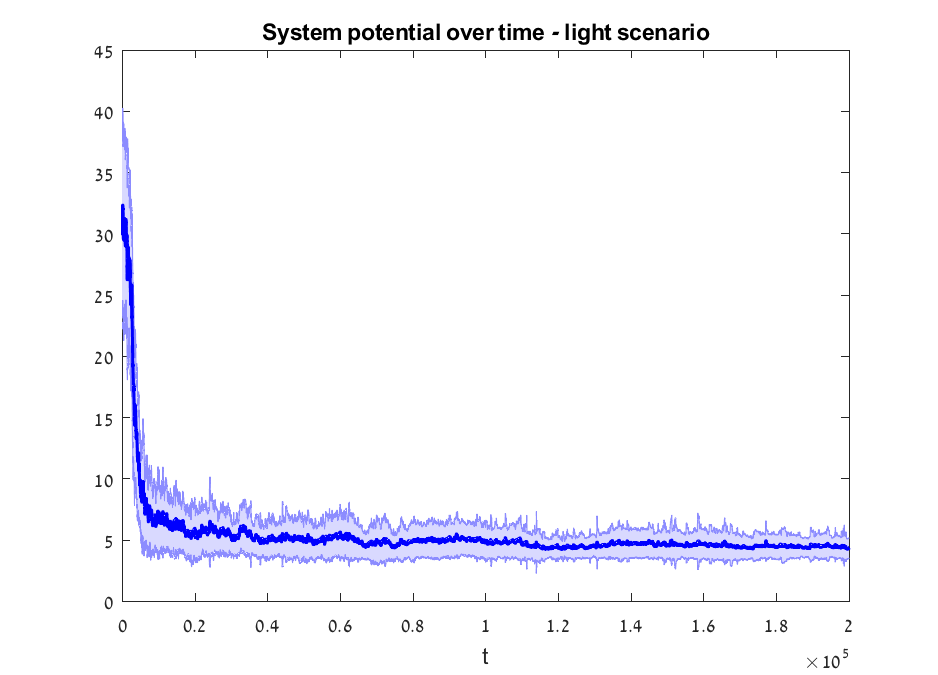}
\caption{Potential - light scenario: K=10, N=7}
\label{fig:Potential1}
\end{figure}

\begin{figure}[!t]
\centering
\includegraphics[width=\columnwidth]{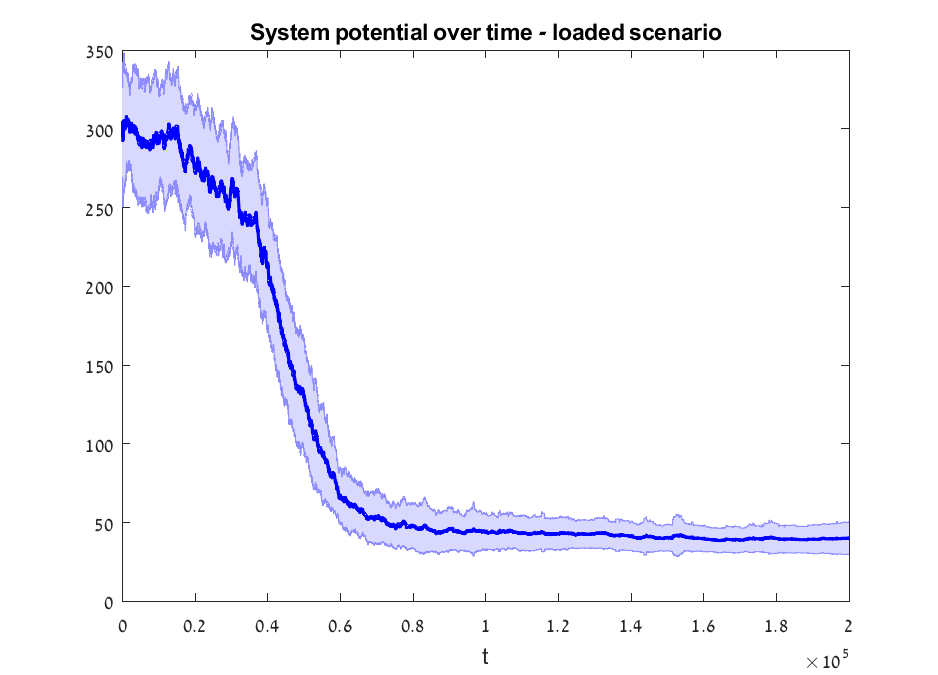}
\caption{Potential - loaded scenario: K=25, N=25}
\label{fig:Potential2}
\end{figure}

\begin{figure}[!t]
\centering
\includegraphics[width=\columnwidth]{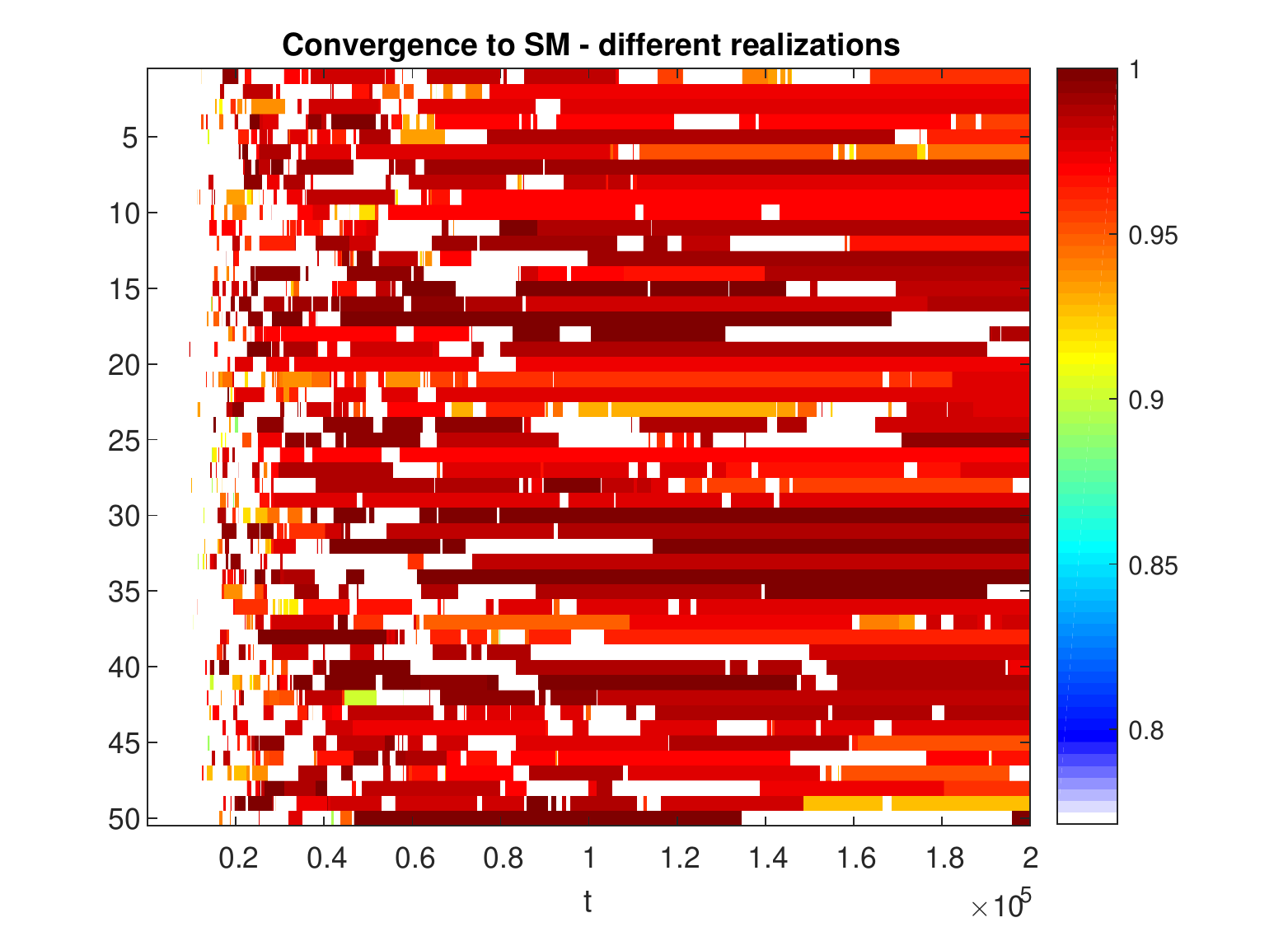}
\caption{Stable configurations: K=15, N=15}
\label{fig:SM}
\end{figure}

Our next result examines the convergence to different stable marriage configurations over time, comparing different realizations of a single setup.
Each line in \figref{fig:SM} represents a single repetition of an experiment with 15 users and 15 channels. The horizontal axis is the time axis, and the values (colors) of the pixels correspond to the ratio between the actual and optimal system-wide rewards. White pixels correspond to time slots during which the system was not in an SMC. This figure shows several interesting phenomena. First, it is quite obvious that the users spend most of their time in high-reward configurations. Convergence to stable (and rewarding) configurations is visible as well; white pixels become scarce as time advances. Finally, the variability between different realizations is demonstrated: for the specific setting of this experiment, 519 different stable realizations exist. The CSM-MAB algorithm doesn't necessarily converge to the same one, but it clearly favours those with high rewards (see colorbar).

\begin{figure}[!t]
\centering
\includegraphics[width=\columnwidth]{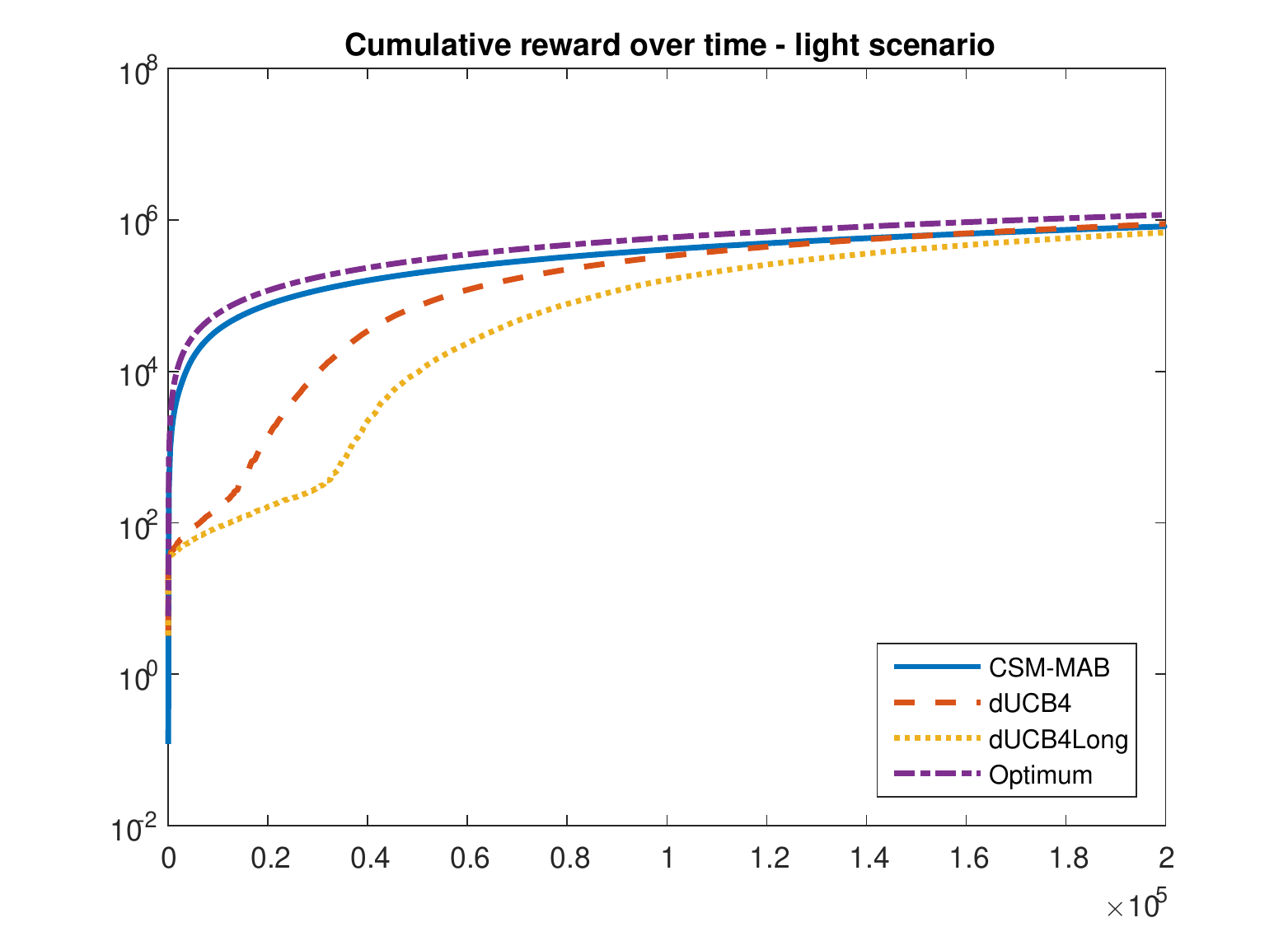}
\caption{Cumulative reward for light load: K=10, N=7}
\label{fig:reward1}
\end{figure}

\begin{figure}[!t]
\centering
\includegraphics[width=\columnwidth]{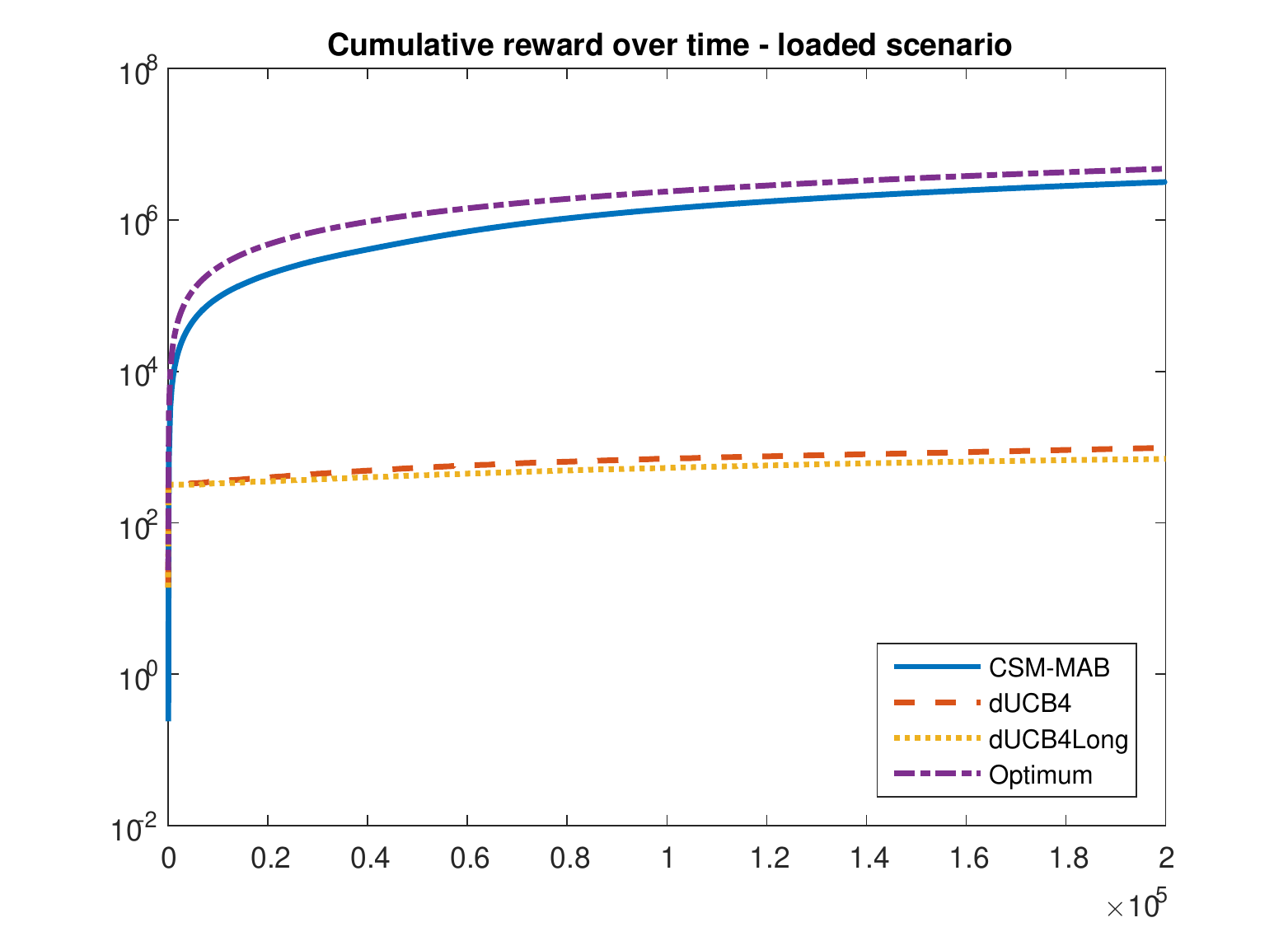}
\caption{Cumulative reward for heavy load: K=25, N=25}
\label{fig:reward2}
\end{figure}

The issue of the reward acquired by users is of great interest in this setup. While our theoretical analysis focuses on stability, we use simulations to complement it empirically and examine the total reward over time. Since users' preferences over channels are driven by the UCB index, based on the channels' rewards, we expect our algorithm to perform reasonably despite not being reward optimal. We compare the performance of our algorithm to that of the dUCB4 algorithm, introduced in \cite{Kalathil2014}. As discussed in \secref{sec:intro}, this algorithm converges to a reward optimal configuration by employing excessive communication in the form of the Bertsekas auction algorithm.
\figref{fig:reward1} and \figref{fig:reward2} present simulation results for light and heavy loads, respectively. Our algorithm (solid blue line) compares well with the system-wide optimal reward (dashed purple) and clearly outperforms both versions of the dUCB4 algorithm (dashed red and dotted orange). The two dUCB4 algorithms differ in the accuracy of the auctioning routine. The ``dUCB4'' version uses 32 bits to encode variables, while the ``dUCB4Long'' version uses 64 bits.

The plots clearly demonstrate what we perceive as the biggest disadvantage of communication between users: its cost. Users applying the CSM-MAB algorithm converge to a ``good'' configuration in terms of reward much faster than those applying the dUCB4 algorithm, since the latter spend a considerable amount of time auctioning. They eventually converge to an optimal configuration, but the time it will take them to reach it may be prohibitively long from a practical point of view. The difference becomes striking as the load on the system increases. Despite running experiments for a very long time, the dUCB4 algorithm did not converge to a favourable configuration in terms of reward in the heavily loaded scenario.

The authors of the dUCB4 algorithm suggest a similar approach in \cite{Nayyar2016}, which includes an exploration phase based on round-robin sampling. Avoiding collisions during this phase requires some form of external coordination (for example, a control channel), which is more extensive than the one we require in this work. Therefore we did not include it in our experiments.

\begin{figure}[!t]
\centering
\includegraphics[width=\columnwidth]{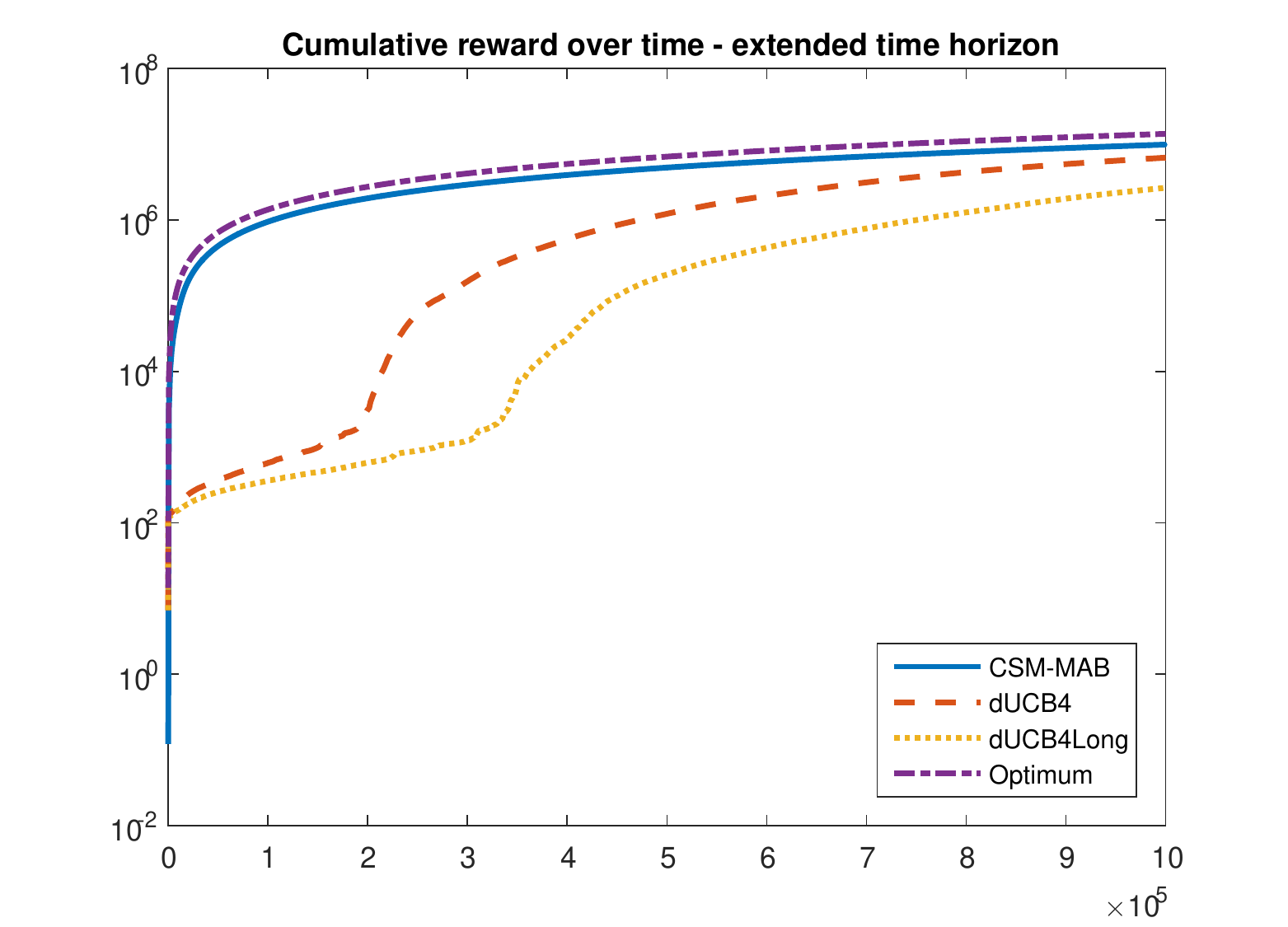}
\caption{Cumulative reward for K=15, N=15, extended time horizon}
\label{fig:reward3}
\end{figure}

Let us examine the price of such communication using a real world example: an average 802.11n WLAN network. We assume the network has a frame size of 2000 bits and bit rate of 25 megabits per second, and analyze a medium-load scenario, with 15 users and 15 channels (see \figref{fig:reward3}).
The reward acquired by the light version of the dUCB4 algorithm comes within one tenth of the optimum after roughly $T=5\cdot10^5$ time slots. Translating this into time results in a start-up phase of 40 seconds: $\frac{5\cdot 10^5\cdot2000}{25\cdot 10^6} = 40 \text{sec}$. The length of this initial phase doubles to well over one minute when 64 bit accuracy is used for the Bertsekas auction. While lighter schemes than the 802.11 can be used, these numbers clearly demonstrate the significant, sometimes impractical, overhead brought on by communication.

\begin{figure}[!t]
\centering
\includegraphics[width=\columnwidth]{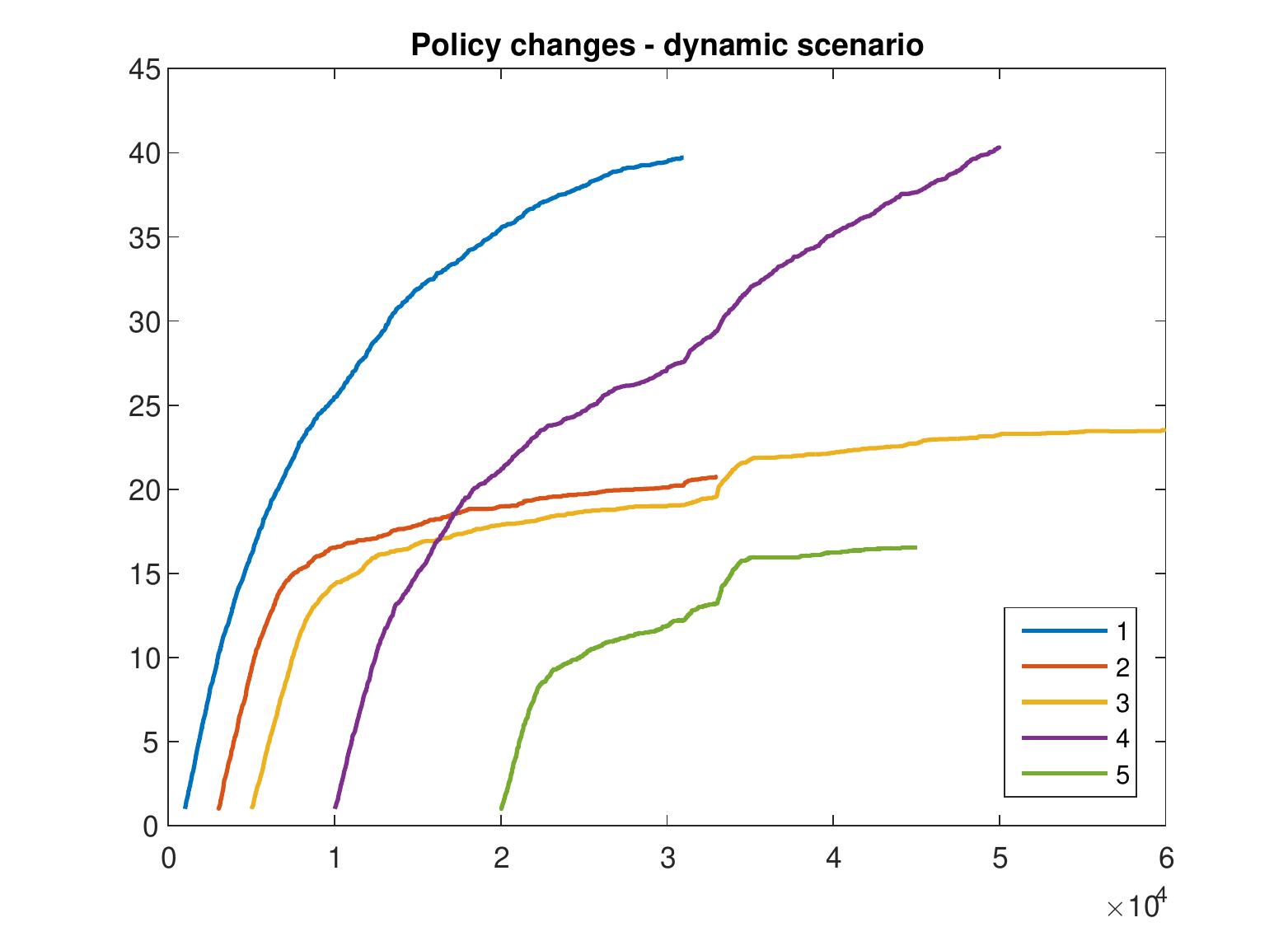}
\caption{Policy changes for dynamic scenario}
\label{fig:policyDynamic}
\end{figure}

Another important contribution of our paper is the D-CSM-MAB algorithm (\secref{sec:D-CSM-MAB}), that is suited for handling a dynamic number of users. We present the results of an experiment where users arrive and leave at different times. The algorithm has no prior knowledge regarding the number of users or when they are expected to arrive or leave. \figref{fig:policyDynamic} displays the empirical average over the cumulative number of policy switches for a network with $K=10$ channels and a variable number of users. The plots are drawn starting with each user's arrival time and are cut off when the user leaves the network. Two interesting phenomena can be observed: first, the rate of policy switches decreases over time, indicating convergence to a stable configuration, similarly to the static scenario. Second, the rate of convergence varies between users, and is also affected by changes in the number of users present. For example, the departure of user 2 (red plot) triggers a change in the actions of the remaining users 3, 4 and 5 (orange, purple and green plots).

\begin{figure}[!t]
\centering
\includegraphics[width=\columnwidth]{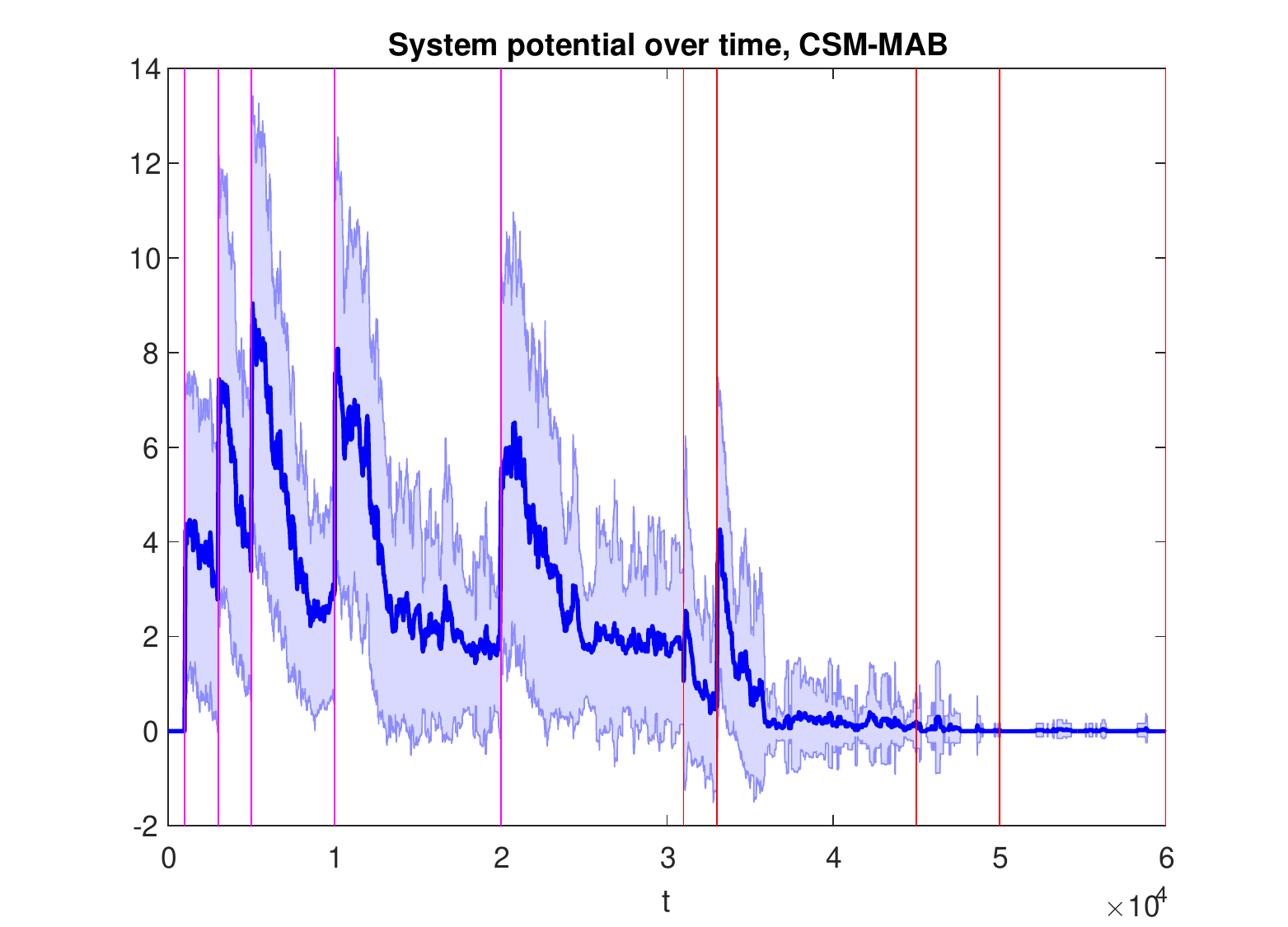}
\caption{Potential for dynamic scenario}
\label{fig:potentialDynamic}
\end{figure}

As mentioned above, the system-wide potential can also be used to examine convergence and stability. \figref{fig:potentialDynamic} shows the changes in potential over time, emphasizing the effect of user arrivals and departures. The vertical lines in the figure represent changes in the number of users: pink lines represent arrivals, while red lines represent departures. Naturally, the manifestation of changes in the number of users is a spike in the system-wide potential: an arriving user has initial high potential since she has yet to learn and choose the best available channel, and a departing user frees a channel the remaining users may have been waiting to sample. The D-CSM-MAB algorithm handles these situations well, as is obvious from the decay of the potential after every change point.


\begin{figure}[!t]
\centering
\includegraphics[width=\columnwidth]{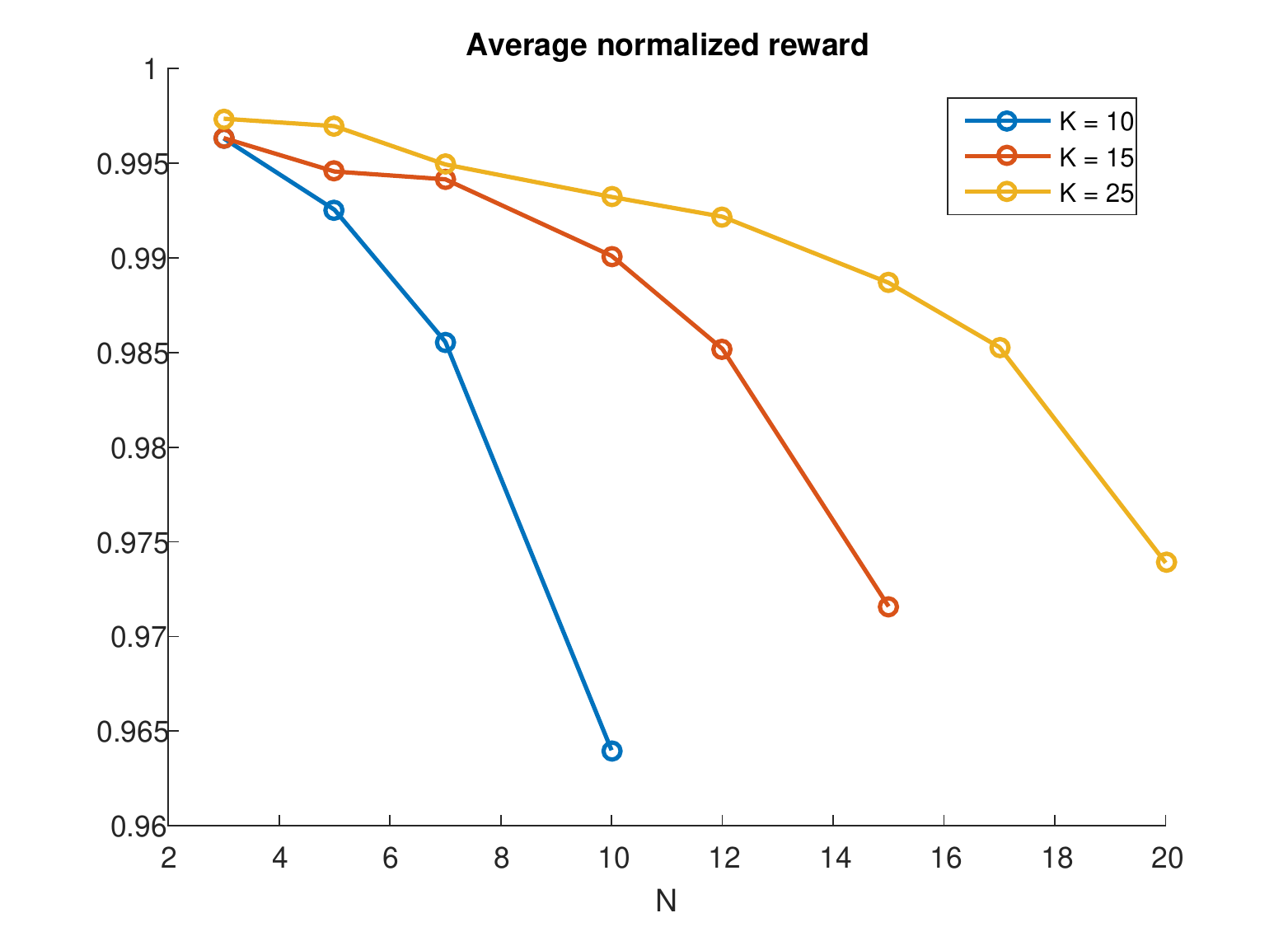}
\caption{Stable configuration reward vs. $K$ and $N$}
\label{fig:rewardKN}
\end{figure}

Our final experiment concerns a fixed number of users once again, and examines the performance of the CSM-MAB algorithm (\secref{sec:CSM-MAB}) over a wide range of setups. The plots in \figref{fig:rewardKN} shows three different network sizes (10, 15, and 25 channels) and several numbers of users, where $N\in \set{3,\ldots,K}$. For each pair of $K$ and $N$ we drew 50 different realizations of reward distributions, and examined the expected reward of the stable configuration the system eventually reached. This value, normalized by the reward of the system-wide optimum, is presented in the figure. For example, with $K=25$ channels and $N=5$ users, the average stable configuration reward is 99.7\% of the optimal reward. Naturally, the steady state reward decreases as the load (i.e., ratio of $N$ to $K$) increases, but even for $N=K$ it stays rather high - over 96\%. The learning problem becomes harder with the increase of $K$, also affecting the stable state reward.

\section{Discussion}\label{sec:conclusion}
Our work addresses the multi-user MAB problem, where users' reward distributions differ and the number of users is unknown.
In an effort to minimize communication between users and still ensure convergence to a favourable configuration, we introduce the CSM-MAB algorithm.
Our approach combines learning with a signalling-based coordination scheme and is guaranteed to converge to an orthogonal stable marriage configuration.

In this paper we present two algorithms. 
The CSM-MAB algorithm deals with a fixed, yet unknown, number of users in the the network, while its dynamic counterpart, the D-CSM-MAB, handles a variable number of users.
We include a theoretical analysis of each of the algorithms, ensuring convergence to an orthogonal SMC in finite time.

The extensive experimental section covers several aspects such as stability, convergence, reward accumulation and comparison to the existing state of the art.
We believe the results of this section demonstrate the practicality of our solution.

We would like to devote a few words to the issue of rogue users.
Our analyses and guarantees depend on the assumption that all users apply the CSM-MAB (or D-CSM-MAB) algorithm.
However, in real networks, non-compliant agents may also be present. Such agents can be either oblivious or adversarial, static or dynamic.
Their effect on performance largely depends on their transmission profile.

For example, an agent that transmits in a single fixed channel and does not transmit during the ``init'' slots (see \figref{fig:frames} and \figref{fig:frames-d}) will cause users to avoid that particular channel, as its empirical average reward will be low due to collisions. Assuming there are enough channels, users will simply learn a configuration that does not involve it.
However, an agent that transmits during the ``init'' slots is likely to impair the process of coordinating an initiator, thus preventing users from swapping channels and moving towards a favorable configuration.
Assuming the ``foreign'' agent is not malicious, she will have no reason to transmit in channels that are already taken, as this will result in collisions for her as well.
She can attempt to drive a transmitting user from an occupied channel, but this will take longer and longer periods of time as the experiment advances, making such a strategy inefficient for the ``foreign'' agent.

While this paper offers a detailed analysis of the problem and the proposed solution, several additional directions are of interest.
First, a dynamic scenario in terms of network resources can be explored. Varying channel characteristics may be addressed by, for example, selecting channels based on their performance in a finite window.
Another direction aims to improve performance by decreasing the amount of time devoted to coordination as time advances, thus increasing channel utilization.
A time dependent scheme can also be of use when dealing with instability caused by users switching between similar channels (e.g., \figref{fig:PolicySwitch1}). 
In this case, introducing switching costs between channels that increase with time is an interesting solution.

\appendices
\section{Proofs of lemmas from \secref{sec:static-analysis}}\label{app:static}
This appendix is devoted to the proofs of the lemmas cited in \secref{sec:static-analysis}.

\subsection{Proof of \lemref{lem:monoPot}}\label{app:static-lem1}
We would like to show that for all values of $t$ for which $t > \alpha\ln t$, the probability that the potential decreases every time it changes is at least $1-4t^{-4}$, where $\alpha = \frac{32K}{\Delta_{\min}^2}$.

Given that a change in potential occurs at time $t$, it is guaranteed to result in a potential decrease if it benefits both users. This will happen if both users' UCB indices, that guide their decisions, are accurate with respect to the true distribution.

Since we condition on a change in potential, denoting an increase in potential by $\Phi_{\text{Inc}}$ and a decrease in potential by $\Phi_{\text{Dec}}$, we have that
\begin{align*}
  \mP{\Phi_{\text{Dec}}}  = 1 - \mP{\Phi_{\text{Inc}}}.
\end{align*}
Let us upper bound $\mP{\Phi_{\text{Inc}}}$ . For a user $n$ switching from arm $j$ to arm $i$ at time $t$, when $\mu_{n,i} < \mu_{n,j}$,
\begin{align*}
  \mP{\Phi_{\text{Inc}}} = \mP{I_{n,i}\paren{t} \geq I_{n,j}\paren{t} \cap
  \mu_{n,i} < \mu_{n,j}},
\end{align*}
where $I_{n,i}\paren{t}$ is user $n$'s UCB index of arm $i$ at time $t$: \begin{align*}
I_{n,k}\paren{t} = \hat{\mu}_{n,k} + \sqrt{\frac{2\ln t}{s_{n,k}}},
\end{align*}
Following the proof of Theorem 1 of \cite{Auer2002a},
\begin{align*}
  \mathbb{P}&\left\{\Phi_{\text{Inc}}\right\}\\
  &= \mP{\hat{\mu}_{n,i}\paren{t} + c_{t,s_{n,i}}
  \geq \hat{\mu}_{n,j}\paren{t} + c_{t,s_{n,j}} \cap \mu_{n,i} < \mu_{n,j}} \\
  &\leq 2t^{-4},
\end{align*}
provided that
\begin{align}\label{eq:condUCB}
  s_{n,i} \geq \frac{8\ln t}{\Delta_{i,j}^2\paren{n}},
\end{align}
where $s_{n,i}$ is the number of times user $n$ sampled arm $i$ up till time $t$ and $\Delta_{i,j}\paren{n} \triangleq \mu_{n,i} - \mu_{n,j}$.
If \eqref{eq:condUCB} does not hold, then the UCB index may ``mislead'' user $n$, causing her to mistakenly favor arm $i$, despite its lower expected reward. Switching from arm $j$ to arm $i$ will result in an increase in potential. However, once she acquires another sample of arm $i$, its index will decrease. In the meantime, the index of arm $j$ will increase due to the passing time, and the indices will ultimately reflect the correct preference, resulting in a potential decrease.

The extreme value for \eqref{eq:condUCB}, i.e., the largest number of required samples, corresponds to the minimal value of $\Delta_{i,j}\paren{n}$.
Let us define:
\begin{align*}
  \Delta_{n} &\triangleq
  \min_{\substack{i,j\in\set{1,\ldots,K} \\ i\neq j}}\sbrk{\mu_{n,i} - \mu_{n,j}}\\
  \Delta_{\min} &\triangleq \min_{n\in\set{1,\ldots,N}}\Delta_{n}
\end{align*}
Thus, when all arms have been sampled at least
\begin{align}\label{eq:s_min}
  s_{\min} \triangleq \frac{8\ln t}{\Delta_{\min}^2}
\end{align}
times, the probability of an increase in potential is very small.

In order to allow for the coordination protocol, users do not gather informative samples in every time slot. Instead, they gather at least $K-2$ samples in each super frame, whose length is $T_{\text{SF}} = 2 + 2\paren{K-1}=2K$.

Therefore, taking into account the fact that the sampling condition in \eqref{eq:s_min} must apply for all arms, the condition on $t$ is
\begin{align}\label{eq:tCond}
  t > K\frac{T_{\text{SF}}}{K-2}s_{\min} = \frac{16K^2}{\paren{K-2}\Delta_{\min}^2}\ln t > \frac{16K}{\Delta_{\min}^2}\ln t.
\end{align}
For all times $t$ for which \eqref{eq:tCond} holds, if a change in potential occurs, it is a decrease, with probability of at least $1-2t^{-4}$.

When we apply this lemma we will use a quantity $t_{\min}$, an upper bound on the minimal $t$ for which \eqref{eq:tCond} holds.
Using an upper bound on the logarithmic function, introduced in \cite{Hardy1952},
\begin{align*}
  \ln x \leq \frac{x^s}{s} \quad \forall x >1,
\end{align*}
with $s = \half$, we have that $\ln t \leq 2\sqrt{t}$.
We use this bound together with \eqref{eq:tCond}:
\begin{align*}
  t = \frac{16K}{\Delta_{\min}^2}\ln t
  > \frac{32K}{\Delta_{\min}^2} \sqrt{t}.
\end{align*}
Our upper bound is therefore
\begin{align*}
  t_{\min} = \paren{\frac{32K}{\Delta_{\min}^2}}^2.
\end{align*}
Since this expression is finite, we may now use it in our proof.

\subsection{Proof of \lemref{lem:initiator}}\label{app:a1-init}
The probability of a specific user becoming the initiator when there are $\ell$ interested users is
\begin{align*}
  P_s\paren{\e,\ell}&\triangleq\cP{\text{specific initiator}}{\ell\text{ interested}}\\
   &= \e\paren{1-\e}^{\paren{\ell-1}} \; \forall \ell\in\set{1,\ldots,N}.
\end{align*}
The probability is minimal when all $N$ users would like to become the initiator, yielding the bound $\e\paren{1-\e}^{N-1}$.

Choosing the value of $\e$, the only parameter in our algorithm, is of some interest. The optimal choice of $\e$, that maximizes the probability of a single initiator emerging when there are $N$ interested users, is $\e = \frac{1}{N}$. However, the total number of users is assumed to be unknown at the user level, and therefore a different value is chosen: $\e = \frac{1}{K}$. Choosing a value smaller than $N$ for the denominator prevents the dependency of $P_s$ on $\ell$ from being monotonous, and so we choose $K$, which is guaranteed to be no less than $N$.

\subsection{Proof of \lemref{lem:finite_time_till_switch}}
A decrease in potential does not occur during a super-frame if at least one of the following occurs:
\begin{itemize}
  \item An initiator does not emerge
  \item An initiator does emerge, but the users' statistics are wrong and therefore a decrease in potential does not occur
\end{itemize}
Formally, denoting the number of decreases in potential in a time interval by $\sbrk{t_1,t_2}$ by $D_{\sbrk{t_1,t_2}}$ and using the bounds of \lemref{lem:monoPot} and \lemref{lem:initiator}:
\begin{align*}
  \mP{D_{\sbrk{t,t+T_{SF}}} = 0} \leq 1-\e\paren{1-\e}^{N-1} + 2t^{-4}.
\end{align*}

\section*{Acknowledgment}

\bibliographystyle{IEEEtran}
\bibliography{journalBib}

\end{document}